\newif\ifpreprint
\preprinttrue
\documentclass{article}
\usepackage{microtype}
\usepackage{graphicx}
\usepackage{booktabs}
\usepackage{amsmath,amssymb}
\usepackage[T1]{fontenc}
\usepackage{hyperref}
\usepackage{xurl}
\usepackage{tikz}
\usetikzlibrary{arrows.meta,positioning,backgrounds,fit}
\usepackage{pgfplots}
\pgfplotsset{compat=1.18}
\usepackage{placeins}
\let\SavedAddcontentsline\addcontentsline
\ifpreprint
  \usepackage[preprint]{icml2026}
\else
  \usepackage{icml2026}
\fi
\ifpreprint\let\addcontentsline\SavedAddcontentsline\fi

\newcommand{\EncodingUTFBits}{24}
\newcommand{\EncodingPayloadBits}{16}
\newcommand{\EncodingPrefixBits}{8}
\newcommand{\EncodingLEBits}{16}
\newcommand{\EncodingUTFFirstBits}{4}
\newcommand{\EncodingLEFirstBits}{8}

\newcommand{\ExampleSentences}{300}
\newcommand{\ExampleTokens}{6}

\newcommand{\ExampleInsideScalar}{3}
\newcommand{\ExampleInsideCluster}{2}
\newcommand{\ExampleSnappedUnits}{1}
\newcommand{\ExampleClusterUnits}{4}
\newcommand{\ExampleBytes}{16}
\newcommand{\ExampleUBESymbols}{11}
\newcommand{\AOneLanguages}{17}
\newcommand{\AOneSites}{2,363,895}
\newcommand{\AOneBoundaries}{1,507,877}
\newcommand{\AOneInside}{856,018}
\newcommand{\AOnePairs}{17,204}
\newcommand{\AOneSteps}{4,000}
\newcommand{\AOneRows}{64}
\newcommand{\AOneHidden}{512}
\newcommand{\AOneSpan}{384}
\newcommand{\AOneTrainMB}{8}
\newcommand{\AOneCells}{18}
\newcommand{\AOneSeeds}{2}
\newcommand{\AOneUTFOne}{8.76}

\newcommand{\AOneUBEOne}{0.06}

\newcommand{\AOneBEOne}{8.62}

\newcommand{\AOneUTFMiss}{9.50}
\newcommand{\AOneUTFFalseCut}{8.01}
\newcommand{\AOneUBEMiss}{0.10}
\newcommand{\AOneUBEFalseCut}{0.02}
\newcommand{\AOneUBEWithin}{0.19}
\newcommand{\AOneUTFWithinMax}{14}
\newcommand{\AOneUTFWithinPositions}{4,077,681}
\newcommand{\AOneUTFTwoPairsLow}{27.2}
\newcommand{\AOneUTFTwoPairsHigh}{28.8}
\newcommand{\AOneUTFTwoCorrectLow}{4,676}
\newcommand{\AOneUTFTwoCorrectHigh}{4,949}
\newcommand{\AOnePairsTwoBytes}{9,313}
\newcommand{\AOnePairsThreeBytes}{7,891}
\newcommand{\AOneGapUBEUTF}{8.69}
\newcommand{\AOneGapUBEUTFLow}{8.63}
\newcommand{\AOneGapUBEUTFHigh}{8.76}
\newcommand{\AOneGapUBEBE}{8.55}
\newcommand{\AOneGapUBEBELow}{8.49}
\newcommand{\AOneGapUBEBEHigh}{8.62}
\newcommand{\AOneGapFifth}{0.137}

\newcommand{\AOneJoinerSites}{5,776}
\newcommand{\AOneJoinerShare}{0.24}
\newcommand{\AOneFullLow}{0.16}
\newcommand{\AOneFullHigh}{0.20}
\newcommand{\AOneFullOtherMax}{0.012}
\newcommand{\AOneUBEOneOther}{0.049}

\newcommand{\AOneUTFOneOther}{8.60}

\newcommand{\AOneScripts}{13}
\newcommand{\CJKCells}{36}
\newcommand{\CJKSeeds}{3}
\newcommand{\CJKSentences}{1,012}
\newcommand{\CJKParameters}{592,512}
\newcommand{\CJKHidden}{256}
\newcommand{\CJKEmbed}{128}
\newcommand{\CJKBatch}{32}
\newcommand{\CJKSpan}{128}
\newcommand{\CJKSteps}{2,000}
\newcommand{\CJKTrainMB}{8}

\newcommand{\CJKChineseSaving}{32.05}
\newcommand{\CJKChineseSplit}{0.61}

\newcommand{\CJKChineseParentUTF}{159.95}
\newcommand{\CJKChineseParentUBE}{103.50}

\newcommand{\CJKChineseParentReduction}{35.29}
\newcommand{\CJKChineseParentBEReduction}{14.63}
\newcommand{\CJKKoreanSaving}{29.10}
\newcommand{\CJKKoreanSplit}{4.74}

\newcommand{\CJKKoreanParentUTF}{54.44}
\newcommand{\CJKKoreanParentUBE}{14.75}

\newcommand{\CJKKoreanParentReduction}{72.91}
\newcommand{\CJKKoreanParentBEReduction}{41.85}
\newcommand{\CJKSpacingRuns}{18}
\newcommand{\CJKChineseWhitespaceShare}{2.44}
\newcommand{\CJKChineseAwayUTF}{17.11}
\newcommand{\CJKChineseAwayUBE}{10.93}

\newcommand{\CJKKoreanWhitespaceShare}{21.86}
\newcommand{\CJKKoreanAwayUTF}{9.94}
\newcommand{\CJKKoreanAwayUBE}{2.69}

\newcommand{\XThreeCells}{18}
\newcommand{\XThreeKoreanCross}{500}
\newcommand{\XThreeChineseCross}{750}
\newcommand{\XThreeCheckpointEvery}{250}
\newcommand{\XThreeKoreanCrossShare}{25}
\newcommand{\XThreeChineseCrossShare}{37.5}

\newcommand{\XOneCells}{144}
\newcommand{\XOneBootstrapDraws}{2,000}
\newcommand{\XOneMinReduction}{67.32}
\newcommand{\XOneMinLanguage}{Japanese}
\newcommand{\XOneMaxReduction}{99.55}
\newcommand{\XOneMaxLanguage}{Hindi}
\newcommand{\XOneBEMaxGap}{3.0}
\newcommand{\XOneMinBEShare}{97}
\newcommand{\XOneJapaneseUTF}{118.7}
\newcommand{\XOneJapaneseBE}{109.8}
\newcommand{\XOneJapaneseUBE}{38.8}

\newcommand{\XOneMinGapClosed}{92}

\newcommand{\AuditLanguages}{22}
\newcommand{\FocalLanguages}{18}
\newcommand{\ProbeLanguages}{17}
\newcommand{\AuditSentences}{22,264}
\newcommand{\SentencesPerLanguage}{1,012}
\newcommand{\ProbeItems}{3,400}
\newcommand{\ProbeItemsPerLanguage}{200}
\newcommand{\SourceSentences}{200}
\newcommand{\BootstrapDraws}{4,000}
\newcommand{\MatchedItems}{850}
\newcommand{\SubsetPerLanguage}{50}
\newcommand{\MinCutRetention}{97.07}
\newcommand{\MaxInternalChange}{1.54}
\newcommand{\ParentMultiSplit}{99.89}
\newcommand{\BwenMultiSplit}{99.49}
\newcommand{\AblatedMultiSplit}{66.18}
\newcommand{\MaxCoarsening}{15.6}
\newcommand{\KnifeEdges}{8}
\newcommand{\InteriorPositions}{7,407,006}
\newcommand{\ThresholdBand}{10^{-4}}

\newcommand{\ParentEnglish}{0.2093}
\newcommand{\ParentRatio}{6.10}
\newcommand{\BwenDensity}{1.2743}
\newcommand{\BwenEnglish}{0.2057}
\newcommand{\BwenRatio}{6.19}
\newcommand{\SnappedParentDensity}{0.4705}
\newcommand{\SnappedParentEnglish}{0.2093}
\newcommand{\SnappedParentRatio}{2.25}

\newcommand{\RemovedUnits}{63.15}
\newcommand{\UBESaving}{31.6}
\newcommand{\OdiaTokens}{267,893}
\newcommand{\OdiaSymbols}{248,774}
\newcommand{\OdiaClusters}{82,869}
\newcommand{\EmbeddingMillions}{621.3}
\newcommand{\EmbeddingShare}{7.47}
\newcommand{\AlphabetMillions}{4.19}
\newcommand{\AlphabetShare}{0.0505}
\newcommand{\HiddenWidth}{4,096}
\newcommand{\ParentRows}{151,680}
\newcommand{\BwenParameters}{8,313,169,056}
\newcommand{\InvalidPieces}{1,448}
\newcommand{\VocabularyPieces}{151,643}
\newcommand{\PatchTokenRatio}{0.9977}
\newcommand{\PatchTokenCorrelation}{0.9998}
\newcommand{\WithinFivePercent}{97.4}
\newcommand{\PatchRatioRange}{0.9710--1.0213}
\newcommand{\PairTwoPatchTokenRatio}{0.9988}
\newcommand{\PairTwoPatchTokenCorrelation}{0.9999}
\newcommand{\PairTwoWithinFive}{99.6}
\newcommand{\UnstableItems}{1,786}
\newcommand{\InstabilityFull}{52.5}
\newcommand{\PrefixCorruptions}{6,800}
\newcommand{\PrefixFlagChanges}{0}
\newcommand{\PairOneDelta}{+0.0177}
\newcommand{\PairOneCI}{[+0.0074, +0.0283]}
\newcommand{\PairOneItemDelta}{\textminus{}0.0201}
\newcommand{\PairOneItemCI}{[\textminus{}0.0287, \textminus{}0.0118]}
\newcommand{\PairTwoDelta}{+0.0064}
\newcommand{\PairTwoCI}{[\textminus{}0.0080, +0.0214]}
\newcommand{\PairTwoItemDelta}{\textminus{}0.0048}
\newcommand{\PairTwoItemCI}{[\textminus{}0.0164, +0.0061]}
\newcommand{\SignAgreement}{8}
\newcommand{\CrossPairAll}{+0.408}
\newcommand{\CrossPairOmitMyanmar}{\textminus{}0.219}
\newcommand{\OmitThaiDelta}{+0.0091}
\newcommand{\CheckpointAgreement}{842}
\newcommand{\CheckpointSizeRatio}{5.2}
\newcommand{\FlagCorrelation}{+0.9982}
\newcommand{\InstabilityOneB}{38.6}
\newcommand{\InstabilitySevenB}{38.4}
\newcommand{\BppAssociationRange}{+0.620 to +0.764}
\newcommand{\ShanBytesPerPatch}{1.284}
\newcommand{\ShanInstability}{66}
\newcommand{\OdiaBytesPerPatch}{1.075}
\newcommand{\OdiaInstability}{16}
\newcommand{\BOSInflation}{74.2}
\newcommand{\BOSMultiplier}{11.0}
\newcommand{\BOSItems}{51}
\newcommand{\BatchMaxBits}{12.32}
\newcommand{\BatchPathChanges}{10}
\newcommand{\BatchVariants}{306}
\newcommand{\BatchOneBMaxBits}{41.28}
\newcommand{\BatchOneBPathChanges}{11}
\newcommand{\BatchOneBVariants}{2,550}
\newcommand{\BatchSevenBMaxBits}{20.88}
\newcommand{\BatchSevenBPathChanges}{29}
\newcommand{\BatchSevenBVariants}{2,550}
\newcommand{\FlagLanguageMean}{13.0}
\newcommand{\BolmoCuteMillions}{75}
\newcommand{\BolmoMixBillions}{172}
\newcommand{\BolmoCutePercent}{0.04}

\newcommand{\papertitle}{The Right Byte First:\texorpdfstring{\\}{ }How UTF-16 Helps Multilingual Boundary Prediction}
\icmltitlerunning{The Right Byte First}
\hypersetup{pdftitle={\papertitle},
  pdfsubject={Byte order and limited-lookahead multilingual boundary prediction with UTF-16},
  pdfkeywords={byte-level language models, UTF-16, multilingual boundary prediction, grapheme clusters}}
\newcommand{\ain}{A_{\mathrm{in}}}
\newcommand{\bin}{B_{\mathrm{in}}}
\newcommand{\modelid}[1]{{%
  \expandafter\def\expandafter\UrlOrds\expandafter{\UrlOrds\do\.}%
  \def\UrlBreaks{\do\/}\def\UrlBigBreaks{}\nolinkurl{#1}}}
\begin{document}
\twocolumn[
\icmltitle{\papertitle}
\ifpreprint
  \begin{icmlauthorlist}
    \icmlauthor{Hyunsik Kim}{}
    \par\vspace{0.2em}
    {\normalfont Samsung Research}
    \par\vspace{0.15em}
    {\normalfont\small\href{mailto:hyunsik777.kim@samsung.com}{\nolinkurl{hyunsik777.kim@samsung.com}}}
  \end{icmlauthorlist}
\else
  \begin{icmlauthorlist}
    \icmlauthor{Anonymous Authors}{}
  \end{icmlauthorlist}
  \icmlaffiliation{anon}{Anonymous Institution}
  \icmlcorrespondingauthor{Anonymous Author}{anon.email@domain.com}
\fi
\icmlkeywords{Byte-level language models, UTF-16, multilingual boundary prediction, grapheme clusters}
\vskip 0.3in
]
\ifpreprint
  \makeatletter
  \global\icml@noticeprintedtrue
  {\let\thefootnote\relax\footnotetext{\Notice@String}}
  \makeatother
\else
  \printAffiliationsAndNotice{}
\fi
\begin{abstract}
Many hierarchical byte-level language models group bytes into patches using a learned boundary predictor whose view of the next codepoint depends on encoding and lookahead.
In \mbox{UTF-8}, codepoints of many scripts take three bytes: \EncodingPrefixBits{} of \EncodingUTFBits{} bits mark length and continuation, and the first byte carries only \EncodingUTFFirstBits{} high codepoint bits.
After the same left context, a mark extending a grapheme cluster and a letter starting the next can thus share their first byte.
Universal Byte-Level Encoding (UBE) keeps one- and two-byte \mbox{UTF-8} encodings but writes these codepoints as two \mbox{UTF-16} bytes, low byte first, so its first symbol carries \EncodingLEFirstBits{} low codepoint bits that can distinguish such continuations.
We train small causal predictors on matched FineWeb-2 text and evaluate on FLORES-200 under \mbox{UTF-8}, UBE, and UBE's same-length big-endian variant.
At one-symbol lookahead, balanced grapheme-boundary error at codepoint boundaries across \AOneLanguages{} languages in \AOneScripts{} Brahmic scripts is \AOneUTFOne\% with \mbox{UTF-8}, \AOneBEOne\% with the big-endian variant, and \AOneUBEOne\% with UBE.
When the target instead retains a parent tokenizer's subword cuts at grapheme boundaries, UBE lowers the mean error relative to \mbox{UTF-8} at the same lookahead by \CJKKoreanParentReduction\% in Korean, \CJKChineseParentReduction\% in Chinese, and \mbox{\XOneMinReduction--\XOneMaxReduction\%} in eight more scripts.
In the seven of them whose main Unicode block fits on one 256-codepoint page, the big-endian variant stays within \XOneBEMaxGap\% of \mbox{UTF-8}, so the gain follows byte order, not length.
In Korean and Chinese, UBE's mean error falls below \mbox{UTF-8}'s final mean error after only \XThreeKoreanCrossShare\% and \XThreeChineseCrossShare\% of training.
Which byte comes first thus affects how much lookahead a byte-level boundary predictor needs.
\end{abstract}

\section{Introduction}
In hierarchical byte-level language models with learned patching, a boundary predictor groups the input into patches for the global transformer.
The character distinctions needed to predict these units depend on the writing system and the segmentation target.
Consider a grapheme cluster, Unicode's approximation to a written character, which may combine a base with dependent marks.
After identical left context, one-byte lookahead at a codepoint boundary in UTF-8 can reveal the same byte for a mark extending the current cluster and a letter starting the next.
The required decisions then remain indistinguishable even when the target segmentation is unambiguous.
Byte-level models can process text in any language out of the box \citep{byt5}; at limited lookahead, however, the encoding does not expose every script's distinctions equally.
When clusters consist of single codepoints, the task changes: a subword boundary predictor must select which inter-character boundaries to retain.
These tasks require different distinctions, motivating us to study encoding choice in relation to both the script and the boundary target.

\begin{figure*}[t]
\centering
\resizebox{\textwidth}{!}{
\begin{tikzpicture}[line width=0.6pt]
\pgfplotsset{overview/.style={
  scale only axis,width=4.3cm,height=3.2cm,
  xmin=-.74,xmax=1.86,ymin=0,
  xlabel={Lookahead},line width=0.6pt,
  xtick={0,1.25},xticklabels={{One\\symbol},{Whole next\\scalar}},xticklabel style={align=center},
  axis lines*=left,axis line style={line width=0.6pt},tick style={black,line width=0.6pt},
  ymajorgrids=true,grid style={black!60,line width=0.6pt},
  every axis plot/.append style={line width=0.6pt},
  tick label style={font=\small},label style={font=\small},
  title style={font=\small,align=center},xtick style={draw=none},
  enlarge x limits=false,clip=false},
  obar/.style={ybar,bar width=11.5pt,bar shift=0pt,area legend},
  olabel/.style={only marks,mark=none,forget plot,nodes near coords,point meta=explicit symbolic,
    every node near coord/.append style={font=\fontsize{7}{8}\selectfont,text=black,anchor=south,yshift=1.2pt,fill=white,inner sep=0.8pt}},
  outf/.style={obar,fill=blue!25,draw=blue!65!black},
  obe/.style={obar,fill=gray!35,draw=gray!65!black},
  oube/.style={obar,fill=violet!35,draw=violet!75!black},
  oseed/.style={only marks,mark=*,mark size=1.05pt,black,mark options={fill=white}}}
\begin{axis}[overview,name=a,ymax=10,ytick={0,2,4,6,8,10},
  title={(a) Every cluster boundary, \AOneLanguages{} languages},
  ylabel={Balanced error (\%)}]
\addplot[outf] table[x=x,y expr={\thisrow{enc}==0 ? \thisrow{mean} : nan}]{generated/overview_clusters.dat};
\addplot[obe] table[x=x,y expr={\thisrow{enc}==1 ? \thisrow{mean} : nan}]{generated/overview_clusters.dat};
\addplot[oube] table[x=x,y expr={\thisrow{enc}==2 ? \thisrow{mean} : nan}]{generated/overview_clusters.dat};
\addplot[olabel] table[x=x,y expr={\thisrow{group}==0 ? \thisrow{top} : nan},meta=label]{generated/overview_clusters.dat};
\addplot[oseed] table[x expr={\thisrow{x}-.06},y=seed0]{generated/overview_clusters.dat};
\addplot[oseed] table[x expr={\thisrow{x}+.06},y=seed1]{generated/overview_clusters.dat};
\end{axis}
\begin{axis}[overview,name=b,at={(a.south east)},xshift=1.38cm,anchor=south west,
  ymax=65,ytick={0,20,40,60},
  title={(b) Retained parent cuts, Korean},
  ylabel={Errors per 1,000 scalars},
  legend columns=-1,legend style={at={(0.5,1.24)},anchor=south,draw=none,
    font=\small,/tikz/every even column/.append style={column sep=7pt}}]
\addplot[outf] table[x=x,y expr={\thisrow{enc}==0 ? \thisrow{mean} : nan}]{generated/overview_korean.dat};
\addlegendentry{UTF-8}
\addplot[obe] table[x=x,y expr={\thisrow{enc}==1 ? \thisrow{mean} : nan}]{generated/overview_korean.dat};
\addlegendentry{UBE-BE (UBE's length, big-endian)}
\addplot[oube] table[x=x,y expr={\thisrow{enc}==2 ? \thisrow{mean} : nan}]{generated/overview_korean.dat};
\addlegendentry{UBE}
\addplot[olabel] table[x=x,y expr={\thisrow{group}==0 ? \thisrow{top} : nan},meta=label]{generated/overview_korean.dat};
\addplot[oseed] table[x expr={\thisrow{x}-.08},y=seed0]{generated/overview_korean.dat};
\addlegendentry{Seeds}
\addplot[oseed,forget plot] table[x=x,y=seed1]{generated/overview_korean.dat};
\addplot[oseed,forget plot] table[x expr={\thisrow{x}+.08},y=seed2]{generated/overview_korean.dat};
\end{axis}
\begin{axis}[overview,name=c,at={(b.south east)},xshift=1.38cm,anchor=south west,
  ymax=180,ytick={0,40,80,120,160},
  title={(c) Retained parent cuts, Chinese},
  ylabel={Errors per 1,000 scalars}]
\addplot[outf] table[x=x,y expr={\thisrow{enc}==0 ? \thisrow{mean} : nan}]{generated/overview_chinese.dat};
\addplot[obe] table[x=x,y expr={\thisrow{enc}==1 ? \thisrow{mean} : nan}]{generated/overview_chinese.dat};
\addplot[oube] table[x=x,y expr={\thisrow{enc}==2 ? \thisrow{mean} : nan}]{generated/overview_chinese.dat};
\addplot[olabel] table[x=x,y expr={\thisrow{group}==0 ? \thisrow{top} : nan},meta=label]{generated/overview_chinese.dat};
\addplot[oseed] table[x expr={\thisrow{x}-.08},y=seed0]{generated/overview_chinese.dat};
\addplot[oseed] table[x=x,y=seed1]{generated/overview_chinese.dat};
\addplot[oseed] table[x expr={\thisrow{x}+.08},y=seed2]{generated/overview_chinese.dat};
\end{axis}
\end{tikzpicture}}
\caption{Boundary errors at one-symbol lookahead ($k=1$) and with the whole next scalar visible ($k=3$ for UTF-8; $k=2$ for UBE and UBE-BE). (a)~Every-cluster target on the \AOneLanguages-language FLORES-200 panel: balanced error at scalar ends, the mean of the missed-boundary and false-cut rates; within-scalar cuts are reported separately (Section~\ref{sec:lookaheadprobe}). (b,~c)~Retained parent-tokenizer cuts in Korean and Chinese, where every evaluated cluster is one scalar: errors per 1,000 scalars, including within-scalar false cuts. Bars are means over \AOneSeeds{} seeds in (a) and \CJKSeeds{} in (b,~c); dots show seeds.}
\label{fig:overview}
\end{figure*}
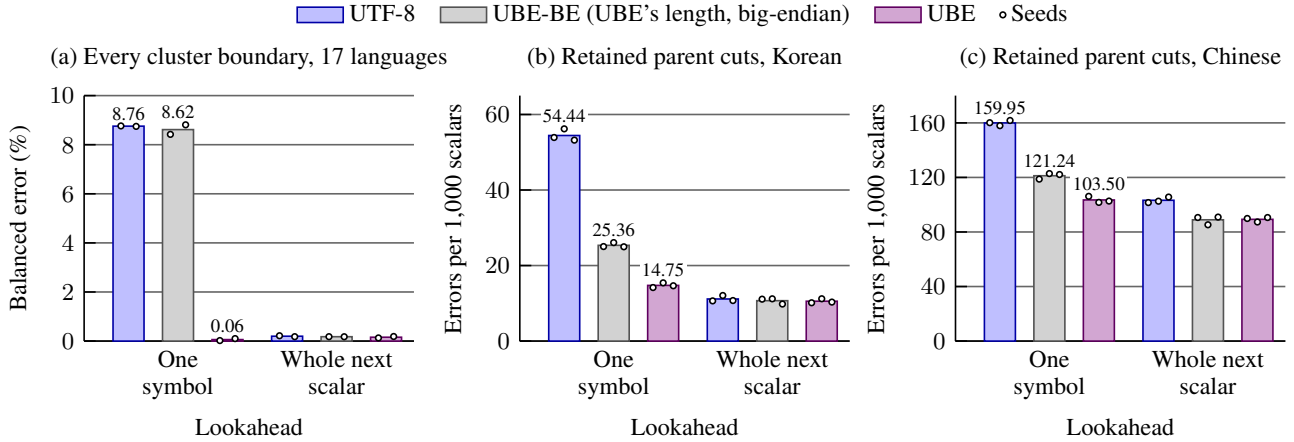

The encoding affects both how densely codepoint bits are packed and which bits reach the predictor first.
For a Unicode scalar (a codepoint other than a surrogate) that takes three UTF-8 bytes, the pattern \texttt{1110xxxx 10xxxxxx 10xxxxxx} carries \EncodingPayloadBits{} codepoint bits and \EncodingPrefixBits{} length/continuation bits \citep{unicode2025}.
The same scalar occupies \EncodingLEBits{} bits in UTF-16 without those markers, so UTF-8 is inherently less compact for such scalars.
We use the existing Universal Byte-Level Encoding (UBE; \citealp{ube}), which retains one- and two-byte UTF-8 encodings and uses UTF-16 little-endian (UTF-16LE) for longer ones.
For three-byte UTF-8 scalars, UBE therefore shortens the representation to two symbols and presents \EncodingLEFirstBits{} low codepoint bits in the first symbol, where UTF-8 presents only \EncodingUTFFirstBits{} high bits.
Within a known 256-codepoint page, that byte identifies the codepoint.
When a letter and a dependent mark occupy different positions on that page, UBE therefore exposes the distinction needed to start or extend a cluster in its first symbol, even if their UTF-8 prefixes coincide.
Codepoints on different pages can still share a low byte.
Figure~\ref{fig:overview} previews the boundary errors of small predictors trained on each encoding; Figure~\ref{fig:encoding} shows why the first byte can hide or expose the required distinction.
The worked example in Appendix~\ref{app:lookahead} illustrates the cluster fragmentation that motivates our targets.

Our motivating application is \emph{byteification}, the conversion of a pretrained subword language model into a model that reads and emits bytes.
Bolmo surrounds the parent transformer with local modules and trains a boundary predictor on the parent's subword segmentation; the global transformer then operates on the resulting patches \citep{bolmo,bolmonature}.
Byteification thus removes the tokenizer from the model's interface but transfers its segmentation criterion along with the weights: the patches are still learned from the parent's cuts (Section~\ref{sec:retained}).
Bolmo raises the question of what models inherit from the input encoding and suggests alternatives to UTF-8 \citep[Section~8]{bolmo}; \citet{haltiuk} reports nearly inherited Bolmo boundaries and proposes separating boundary placement from language modeling.
We focus on how to construct cluster-aligned targets and expose the distinctions needed to predict them with limited lookahead.
The targets either include every grapheme boundary or retain only the parent-tokenizer cuts that already coincide with one.
We use grapheme clusters because their boundaries are representable in every encoding, position-only cluster targets are invariant to span-preserving edits, and cluster continuations are where encodings differ in what the first symbol reveals (Section~\ref{sec:proposal}); the choice does not presuppose that clusters are better modeling units.

\begin{figure*}[t]
\centering
\includegraphics[width=\textwidth]{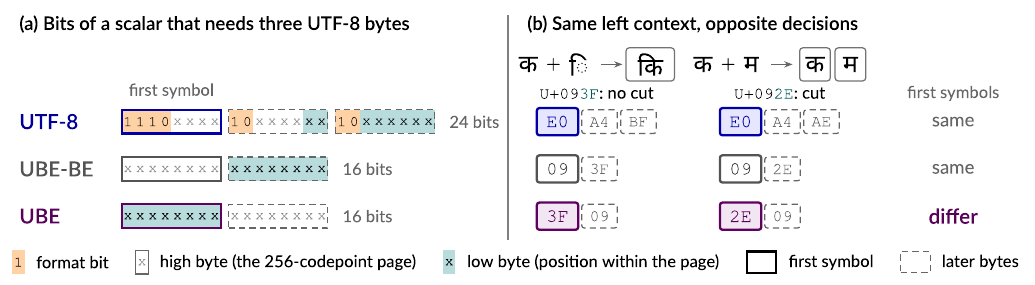}
\caption{How the first symbol can expose the distinction needed for segmentation. (a)~Bit layout of a scalar that needs three UTF-8 bytes (\texttt{x}: codepoint bit) under UTF-8 and UTF-16; UBE writes the low byte first and UBE-BE the high byte first. (b)~After the same Devanagari base, vowel sign I extends the cluster and letter MA starts a new one (outlined: target clusters; dotted circle: isolated mark). Both scalars share the high byte \texttt{09}, so their first symbols coincide under UTF-8 and UBE-BE and differ only under UBE (P5, Section~\ref{sec:proposal}). Byte values are hexadecimal; one-symbol lookahead sees only the solid first symbol.}
\label{fig:encoding}
\end{figure*}

We train small causal boundary predictors on matched FineWeb-2 text \citep{fineweb2} under UTF-8, UBE, and UBE-BE, a big-endian variant with UBE's routing and stream length.
UBE differs from UTF-8 in both stream length and byte order; UBE-BE keeps UBE's length but reverses the bytes within each UTF-16 code unit, allowing a byte-order comparison at fixed stream length.
For the every-cluster target on the \AOneLanguages-language FLORES-200 panel \citep{nllb}, one-symbol lookahead yields \AOneUTFOne\% balanced error at scalar ends with UTF-8 and \AOneUBEOne\% with UBE.
UBE-BE gives \AOneBEOne\%, implicating byte order beyond stream shortening alone.
False cuts inside scalars are measured separately; a rule-based mask can remove them without changing any decision at scalar ends (Section~\ref{sec:lookaheadprobe}).
On constructed pairs with identical left context whose next scalars lie on the same 256-codepoint page but require opposite decisions, the trained UBE predictors answer all \AOnePairs{} pairs correctly at one-symbol lookahead in each seed; UTF-8 and UBE-BE cannot, because both continuations begin with the same symbol.

Korean and Chinese provide a complementary test: every evaluated cluster is a single scalar, so the retained-parent task selects among inter-character boundaries.
In this separate extension, UBE reduces total retained-parent errors relative to UTF-8 by \CJKKoreanParentReduction\% and \CJKChineseParentReduction\%, respectively, at one-symbol lookahead.
These totals include false cuts inside scalars, and every seed favors UBE over both UTF-8 and UBE-BE in each language.
Grapheme boundaries here coincide with scalar ends, which every encoding exposes, so the gain concerns subword selection (Section~\ref{sec:cjk}).
In eight more scripts, UBE reduces retained-parent errors by \mbox{\XOneMinReduction--\XOneMaxReduction\%} at one-symbol lookahead, and where a script's main Unicode block lies on one 256-codepoint page, the same-length UBE-BE control stays close to UTF-8 (Section~\ref{sec:scripts}).

An audit of released conversions motivates the target construction.
Bwen-8B, a released Bolmo-style conversion of Qwen3-8B-Base \citep{bolmo,bolmonature}, retains at least \MinCutRetention\% of its parent's interior cuts in each of \FocalLanguages{} selected high-fragmentation languages, and its cluster-internal cut count stays within \MaxInternalChange\% of the parent's.
Deleting cluster-internal cuts yields supervision that respects cluster boundaries and is representable under UTF-8, UTF-16, and UBE.
On the \ProbeLanguages-language panel, this deletion reduces target units per normalized codepoint from \ParentRatio$\times$ to \SnappedParentRatio$\times$ the English rate.
Complementary corruption and scoring analyses characterize the released segmentation and establish how to compare a conversion with its parent.

Our contributions are:
\begin{itemize}
\item An analysis of the information available for boundary prediction at limited lookahead, tested with trained UTF-8, UBE, and UBE-BE predictors on grapheme and retained subword targets.
\item Cluster-aligned boundary targets with explicit integrity, representability, and invariance conditions, separating target construction from learned boundary decisions.
\item An audit of grapheme fragmentation in released conversions, with sequence-length accounting, corruption diagnostics, and a scoring protocol for parent/conversion comparisons.
\end{itemize}
All model diagnostics use released checkpoints and the FLORES-200 devtest split; the boundary predictors train on FineWeb-2 and are evaluated on FLORES-200.
P1\nobreakdash--P4 (Section~\ref{sec:proposal}) are properties of the targets that hold by construction; P5 is an information limit for causal predictors with limited lookahead; Section~\ref{sec:lookaheadprobe} measures trained probes and Section~\ref{sec:released} measures released checkpoints.

\section{Background: Encodings and Byteification}
\label{sec:background}
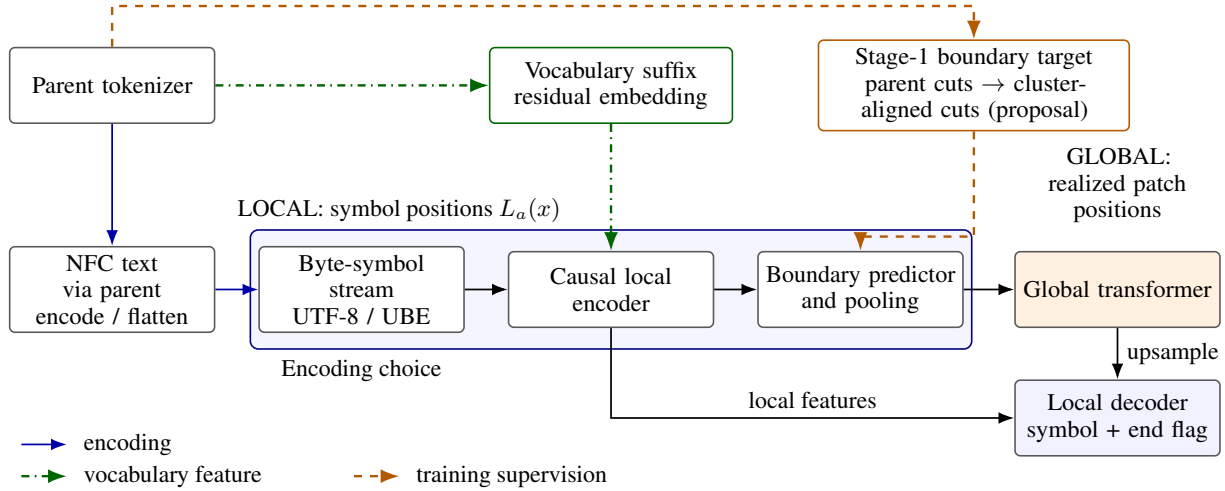
\begin{figure*}[t]
\centering
\begin{tikzpicture}[
  font=\small,>=Latex,line width=0.6pt,
  box/.style={draw=black!65,fill=white,rounded corners=2pt,align=center,
              minimum height=10mm,text width=25mm,inner sep=3pt},
  flow/.style={->,line width=.65pt},
  encode/.style={flow,draw=blue!65!black},
  feature/.style={flow,draw=green!40!black,dash dot,line width=.9pt},
  targetflow/.style={flow,draw=orange!70!black,dashed,line width=.9pt}]

\node[box] (parent) at (0,4.1) {Parent tokenizer};
\node[box] (text) at (0,1.4) {NFC text via parent\par encode / flatten};
\node[box] (symbols) at (3.3,1.4) {Byte-symbol stream\par UTF-8 / UBE};
\node[box] (encoder) at (6.6,1.4) {Causal local\par encoder};
\node[box] (boundary) at (9.9,1.4) {Boundary predictor\par and pooling};
\node[box,fill=orange!12] (global) at (13.3,1.4) {Global transformer};
\node[box,fill=blue!5] (decoder) at (13.3,-.3) {Local decoder\par symbol + end flag};
\node[box,draw=green!40!black,text width=30mm] (vocab) at (6.6,4.1) {Vocabulary suffix\par residual embedding};
\node[box,draw=orange!70!black,text width=39mm] (target) at (11.4,4.1)
  {Stage-1 boundary target\par parent cuts $\to$ cluster-aligned cuts (proposal)};

\begin{scope}[on background layer]
\node[draw=blue!50!black,fill=blue!4,line width=0.6pt,rounded corners=3pt,
      fit=(symbols)(encoder)(boundary),inner xsep=3pt,inner ysep=6pt] (localregion) {};
\end{scope}
\node[font=\small,anchor=south] at (3.8,2.16) {LOCAL: symbol positions $L_a(x)$};
\node[font=\small,align=center,text width=28mm,anchor=south] at (13.3,2.16) {GLOBAL:\par realized patch\par positions};

\draw[encode] (parent.south) -- (text.north);
\draw[encode] (text.east) -- (symbols.west);
\draw[feature] (parent.east) -- (vocab.west);
\draw[feature] (vocab.south) -- (encoder.north);
\draw[targetflow] (parent.north) -- (0,5.15) -- (11.4,5.15) -- (target.north);
\draw[targetflow] (target.south) -- (11.4,2.1) -- (9.9,2.1) -- (boundary.north);
\draw[flow] (symbols.east) -- (encoder.west);
\draw[flow] (encoder.east) -- (boundary.west);
\draw[flow] (boundary.east) -- (global.west);
\draw[flow] (global.south) -- node[right,font=\footnotesize]{upsample} (decoder.north);
\draw[flow] (encoder.south) -- (6.6,-.3) -- node[above,font=\footnotesize]{local features} (decoder.west);
\node[anchor=north,font=\footnotesize] at (localregion.south -| symbols) {Encoding choice};

\draw[encode] (-1.2,-.65) -- (-.6,-.65);
\node[anchor=west,font=\footnotesize] at (-.5,-.65) {encoding};
\draw[feature] (-1.2,-1.07) -- (-.6,-1.07);
\node[anchor=west,font=\footnotesize] at (-.5,-1.07) {vocabulary feature};
\draw[targetflow] (3.2,-1.07) -- (3.8,-1.07);
\node[anchor=west,font=\footnotesize] at (3.9,-1.07) {training supervision};
\end{tikzpicture}
\caption{Three mechanisms of tokenizer inheritance and two sequence lengths. Blue solid arrows trace input encoding, green dash-dotted arrows vocabulary features, and orange dashed arrows boundary supervision. Cluster alignment changes the supervision target; encoding choice changes the local symbol stream. Transferring the vocabulary features to UBE requires a separate alignment policy.}
\label{fig:channels}
\end{figure*}

\paragraph{Written units and encodings.}
We use \emph{cluster} to mean a UAX~\#29 extended grapheme cluster \citep{uax29}, operationalized by \texttt{regex}'s \texttt{\textbackslash X} on text in Unicode Normalization Form C (NFC).
All codepoints in our normalized text are scalars.
A cluster can contain several scalars, each represented by one or more encoded symbols.
Grapheme clusters do not necessarily coincide with linguistic characters or orthographic syllables.
We therefore distinguish scalar, cluster, and symbol counts throughout.

Word-spacing conventions are distinct from grapheme structure.
Korean regularly separates \emph{eojeols}, spacing units that can contain multiple morphemes, whereas Chinese generally leaves word boundaries unmarked by spaces \citep{morphemematters,hnet}.
In neither language do whitespace boundaries define subword segmentation; Unicode grapheme and word boundaries are also distinct \citep{uax29}.
We therefore distinguish whitespace-adjacent gaps from other scalar gaps when analyzing the Korean/Chinese extension.

UBE uses disjoint UTF-8 and UTF-16LE byte alphabets, with 256 symbols each \citep{ube}.
One- and two-byte UTF-8 scalars stay in UTF-8; three-byte scalars in the Basic Multilingual Plane (BMP) use two UTF-16LE byte symbols; supplementary scalars use four UTF-16LE symbols.
A UBE \emph{symbol} identifies both a byte value and its encoding route; lookahead is measured in symbols.
The encoding changes stream length and byte order but does not itself prevent a learned predictor from cutting inside a scalar or cluster.

\paragraph{Boundary counts.}
We compare segmentations in common text coordinates.
For a nonempty sentence, let $C$ be its set of interior cut offsets in NFC UTF-8 bytes.
We count text units as $N=|C|+1$, excluding Bwen's BOS patch.
Let $\ain$ count cuts strictly inside a cluster and $\bin$ cuts strictly inside a Unicode scalar.
Then $S=\ain-\bin$ counts cluster-internal cuts at scalar boundaries.
These metrics count boundary positions; the fraction of multi-scalar clusters with at least one internal cut separately measures how many clusters are split.

\paragraph{Evaluation text.}
FLORES-200 provides parallel translated sentences \citep{nllb}.
The boundary audit uses \AuditLanguages{} languages with \SentencesPerLanguage{} sentences each, \AuditSentences{} sentences in total: \FocalLanguages{} focal languages, namely those covered by Belebele~\citep{bandarkar2024belebele} among the 30 \mbox{FLORES-200} languages in which the Qwen3 tokenizer places a token boundary inside the largest share of non-ASCII scalars, and English, Russian, Thai, and Simplified Chinese as controls.
The lookahead probe and the cost and corruption analyses use a \ProbeLanguages-language panel in \AOneScripts{} Brahmic scripts: the eligible focal languages plus Thai.
Amharic and Tigrinya cannot supply a substitutable mark pool, while English, Russian, and Chinese fail at least one of the two corruption conditions (mark and base substitution; Section~\ref{sec:behaviour}); those three languages remain reference points in the cost accounting.

\paragraph{Converting a subword model.}
Bolmo combines a local encoder, a boundary predictor, the inherited global transformer, and a local decoder.
The released predictor uses one-byte lookahead during prefill, which its authors found largely sufficient to match subword tokenization \citep{bolmonature}; the decoder jointly predicts a byte and its patch-end flag.
The output alphabet has $(4+256)\times2=520$ identifiers, including special tokens.
During Stage~1, the initial distillation stage, parent cuts supervise the predictor; at inference, the predictor supplies its own boundaries.

The parent tokenizer enters this architecture through three mechanisms (Figure~\ref{fig:channels}).
First, the released input pipeline applies the parent tokenizer and concatenates the bytes of its pieces, preserving its NFC normalization but discarding the piece boundaries from the byte stream.
Second, a reversed byte trie retrieves the longest parent-vocabulary suffix ending at each byte position and adds its embedding to the byte embedding.
These suffix matches need not coincide with tokens in the canonical BPE segmentation.
Third, Stage~1 supervises the boundary predictor with the parent's segmentation while the global transformer remains frozen.
The three mechanisms must be distinguished when interpreting inherited boundaries.

Bolmo's public Qwen conversion recipe\footnote{The code is at {\Urlmuskip=0mu\relax\expandafter\def\expandafter\UrlOrds\expandafter{\UrlOrds\do\.}\def\UrlBreaks{\do\/}\def\UrlBigBreaks{}\href{https://github.com/allenai/bolmo-core/tree/2d8178b16ce73cd40ee31a8973d450defa1cdfef}{\nolinkurl{github.com/allenai/bolmo-core}}} (commit 2d8178b).} weights the local-encoder, cross-entropy, local-decoder, and boundary losses by $[1,1,1,4]$.
In Stage~2, which trains all parameters end to end, Bolmo can optionally supervise merged subsets of the parent's boundaries to raise compression \citep[Section~5.1]{bolmo}.
Although the same implementation provides several boundary-compression modes, its Stage-1 entry point rejects gradual boundary compression.
Changing the Stage-1 boundary target therefore requires adapting the Stage-1 pipeline and its alignment to the teacher.

\section{Boundary Targets and the First Byte}
\label{sec:proposal}
We define targets in text coordinates so that encoding comparisons preserve the segmentation task.
P1\nobreakdash--P4 constrain the target; P5 concerns what a causal predictor can observe when it must reproduce it.
Let $G(x)$ be the set of interior cluster boundaries, represented in scalar coordinates.
For an encoding $a$ that represents each scalar as a contiguous, nonempty sequence of symbols, let $L_a(x)$ denote the stream length and $\phi_a$ map scalar boundaries to symbol boundaries.
A cluster-aligned target is any set $T(x)\subseteq G(x)$, realized in the encoded stream as $\phi_a(T(x))$.
This condition constrains the \emph{supervision}; a learned predictor can still deviate from it.
It accommodates different compression criteria: parent-BPE, entropy-based, or fixed-rate cuts can be restricted to $G(x)$ once their offsets are expressed in common coordinates.
The retained-parent construction below removes only the cuts that violate cluster alignment.

\paragraph{P1: target integrity.}
Every target cut is between clusters, hence between scalars.
Both $\ain$ and $\bin$ are zero for the target by construction.
This property follows from the target definition, independently of token frequency or predictor accuracy.

\paragraph{P2: representability across encodings.}
Each retained scalar boundary has a direct image under $\phi_a$, avoiding any need to remap a cut inside a scalar.
For a nonempty sentence with $K$ clusters,
\begin{equation}
  |T(x)|+1 \leq K(x) \leq L_a(x).
\end{equation}
Every target in this family is therefore representable in the encoded stream, a property that unrestricted parent cuts need not satisfy (Section~\ref{sec:lengths}).
For example, the measured Odia text has \OdiaClusters{} clusters and \OdiaSymbols{} UBE symbols.
UTF-8, UTF-16, and UBE can thus represent the same target segmentation despite their different stream lengths.
Vocabulary-feature transfer and the accuracy of the learned predictor remain separate questions.

\paragraph{P3: retaining the compatible parent cuts.}
Let $C_P$ be the parent cut set in UTF-8 byte coordinates.
The retained-parent target, denoted $T_{\rm ret}$, keeps exactly the cuts that already coincide with cluster boundaries:
\begin{equation}
  \begin{aligned}
  T_{\rm ret}(x)&=\{g\in G(x):\phi_{\rm UTF8}(g)\in C_P(x)\},\\
  N_{\rm ret}&=N_P-\ain(P).
  \end{aligned}
\end{equation}
This is a deletion-only operation: it preserves every compatible parent cut and neither relocates nor inserts a boundary.
Only the surviving cuts require scalar coordinates.
On the \ProbeLanguages-language panel (Table~\ref{tab:lengths}), the resulting target has \SnappedParentDensity{} units per codepoint, compared with \SnappedParentEnglish{} in English, a ratio of \SnappedParentRatio$\times$.
The operation reduces the panel's parent-unit count by \RemovedUnits\% and leaves the English target unchanged.
These are target counts; realized counts depend on the learned predictor.

\paragraph{P4: invariance of position-based targets.}
A target determined solely by cluster-boundary positions is invariant, in scalar coordinates, to the span-preserving corruptions in Section~\ref{sec:behaviour}.
Examples include every cluster boundary and every $m$th cluster boundary from the start.
The corruption constructor verifies equality of the complete cluster-span list; equal UTF-8 width of the substituted scalars also preserves offsets under UTF-8, UTF-16, and UBE.
Content-dependent targets, including the retained-parent target, need not be invariant because the selection among valid cluster boundaries can change.
Even for an invariant target, the learned predictor may respond differently to the edited text.

\paragraph{Costs and interpretation.}
Deleting cuts both aligns and lengthens patches, requiring Stage-1 teacher representations and decoder targets to be aligned to the merged spans.
A comparison that attributes a quality change specifically to cluster alignment must control for the accompanying reduction in patch count.
Appendix~\ref{app:proposalinterpretation} specifies these comparison conditions.

\paragraph{P5: the encoding determines visible boundary information.}
Under the released end-boundary convention, the predictor compares projections of local-encoder states $h_i$ and $h_{i+1}$ to decide whether to cut after interior input position $i$; Appendix~\ref{app:lookahead} specifies the BOS offset and endpoint handling.
The released model satisfies three assumptions: the mLSTM local encoder is causal, vocabulary features use only suffixes ending at the current position, and the zero-temperature threshold is deterministic.
Under these assumptions the decision is a function of the symbols through $i+1$, so two continuations with identical left context and the same first symbol must receive the same decision, regardless of training.

For example, after a Devanagari base, vowel sign I (U+093F) extends the cluster while letter MA (U+092E) starts a new one.
Both begin with UTF-8 byte \texttt{E0}.
A target containing every cluster boundary requires opposite decisions for these continuations, which a deterministic UTF-8 predictor with one-byte lookahead cannot provide.
The same ambiguity occurs for the Bengali, Oriya, Sinhala, Thai, Myanmar, and Khmer pairs in Appendix~\ref{app:lookahead}.
The obstruction applies to targets that require this distinction; a subset of $G(x)$ that omits both candidate cuts avoids it.
Reproducing inherited subword cuts at this lookahead, which the released predictor largely does (Section~\ref{sec:retained}), is therefore a different requirement from recovering every cluster boundary.

UBE represents these scalars with UTF-16LE symbols, exposing the low byte first: \texttt{3F} versus \texttt{2E} in the Devanagari example.
Within a known 256-codepoint page, this symbol identifies the scalar and makes the distinction available at one-symbol lookahead.
Different pages can still collide: U+200D ZERO WIDTH JOINER and U+090D DEVANAGARI LETTER CANDRA E both start with low byte \texttt{0D}.
This collision prevents a UBE predictor from reproducing every cluster boundary under the same causal, one-symbol-lookahead restriction, provided its vocabulary features obey that restriction as well.
The argument concerns prefill segmentation of observed text; during generation, the decoder predicts each byte jointly with its boundary flag.
Left context can make a continuation predictable, but encoding visibility alone does not guarantee that training will recover the desired decision.
The target specifies the required cuts; the encoding determines which distinctions the predictor can observe.

\section{Testing the First-Byte Bottleneck}
\label{sec:lookaheadprobe}
P5 establishes an information limit for specific continuations.
We next test how this limit manifests in trained predictors and natural text.
The impossibility that P5 predicts for UTF-8 and UBE-BE on the pairs below holds for any deterministic predictor with this information budget; the natural-text rates characterize the trained probes.

\paragraph{Design and metrics.}
We train small causal predictors with the every-cluster-boundary target $T(x)=G(x)$ under UTF-8, UBE, and UBE-BE.
UBE-BE preserves UBE's routing and stream length but writes each UTF-16 code unit in big-endian order, placing the page byte first for three-byte UTF-8 scalars.
All predictors use a one-layer LSTM with \AOneHidden{} hidden units, the same embedding-table size, and H-Net's cosine boundary head \citep{hnet} in Bolmo's lookahead form \citep{bolmo}, comparing states $k$ symbols apart (Appendix~\ref{app:aone}).
The probes keep Bolmo's boundary rule but replace its mLSTM encoder and parent-vocabulary suffix features with this LSTM.
For each encoding, we train at $k=1,2,3$ with two seeds, matching FineWeb-2 text spans within each seed \citep{fineweb2}.
For each language, training documents containing one of that language's FLORES-200 devtest sentences are excluded.
Evaluation scores all predictors at the same \AOneSites{} scalar ends of the \AOneLanguages-language panel.
All evaluation scalars lie in the BMP, so two UBE symbols cover each next scalar.
Each run takes \AOneSteps{} steps with \AOneRows{} rows of up to \AOneSpan{} scalars, sampled from the same per-language training documents (the leading documents of each language's first FineWeb-2 shard; Appendix~\ref{app:aone}).
Matching text exposure leaves computational cost unmatched because UTF-8 processes more symbol positions.
The objective gives equal aggregate weight to scalar-end and within-scalar decisions.
Balanced error is the mean of the missed-cluster-boundary rate and the false-cut rate at scalar ends inside clusters.
We pool sites across languages before averaging the two seeds, exclude sentence-final boundaries, and report within-scalar false cuts separately.

\begin{table}[t]
\caption{Balanced error (\%, two-seed means) at FLORES-200 devtest scalar ends with $k$ symbols of lookahead. UBE-BE has UBE's routing and stream length but reverses the byte order within each UTF-16 code unit. Within-scalar errors are reported separately in the text and Appendix~\ref{app:aone}.}
\label{tab:aone}
\centering\small
\begin{tabular}{lrrr}
\toprule
Encoding & $k=1$ & $k=2$ & $k=3$ \\
\midrule
UTF-8 & 8.76 & 6.37 & 0.20 \\
UBE & \textbf{0.06} & 0.16 & 0.19 \\
UBE-BE & 8.62 & 0.18 & 0.20 \\
\bottomrule
\end{tabular}

\end{table}

\paragraph{Natural text.}
With one-symbol lookahead, UTF-8 misses \AOneUTFMiss\% of cluster boundaries and places false cuts at \AOneUTFFalseCut\% of scalar ends inside clusters; the corresponding UBE rates are \AOneUBEMiss\% and \AOneUBEFalseCut\% (Appendix Table~\ref{tab:aonefull}).
The balanced-error differences, UTF-8 minus UBE and UBE-BE minus UBE, are \AOneGapUBEUTF{} and \AOneGapUBEBE{} percentage points (95\% paired source-sentence bootstrap intervals \AOneGapUBEUTFLow--\AOneGapUBEUTFHigh{} and \AOneGapUBEBELow\nobreakdash--\AOneGapUBEBEHigh, conditional on two seeds).
Despite having UBE's stream length, UBE-BE has errors comparable to UTF-8 at one symbol, supporting a role for byte order beyond stream length alone; at two symbols, where UBE-BE sees the whole next scalar and UTF-8 does not, its errors are substantially lower.
Reversing the byte order also changes the encoded history, but with the whole next scalar visible to both, UBE and UBE-BE errors nearly coincide at two symbols (Table~\ref{tab:aone}): the contrast is concentrated at one-symbol lookahead.
UBE has the lowest one-symbol error in every panel language; its largest per-language error, in Khmer, varies across the two seeds (Appendix Table~\ref{tab:aonelanguage}).

\paragraph{Same-context continuation pairs.}
To test the information limit directly, we construct a pair from each evaluation sentence with identical left context and same-page continuations requiring opposite decisions.
A pair is correct only if both decisions are correct.
The trained UBE $k=1$ predictors answer all \AOnePairs{} pairs correctly in each seed.
For UTF-8 and UBE-BE at $k=1$, the shared first symbol makes that outcome impossible for any pair, as P5 predicts.
The pairs test the predicted ambiguity by construction; their frequency does not estimate its prevalence in natural text.

\paragraph{Remaining errors.}
Within-scalar false cuts (\AOneUBEWithin\% of within-scalar sites for UBE at $k=1$) are not an information limit: UBE-BE, with the same symbol ranges, makes few, and under all three encodings whether a position ends a scalar is a deterministic function of the encoded prefix, so a rule-based mask can remove these cuts without changing any decision at scalar ends.
Cluster-internal decisions at scalar ends cannot be settled this way and are the subject of P5.
Configurations that see the whole next scalar still have \AOneFullLow--\AOneFullHigh\% balanced error at scalar ends; a post hoc analysis localizes these errors primarily to positions adjacent to zero-width joiner and non-joiner characters (Appendix~\ref{app:aone}).
Away from those sites their maximum error is \AOneFullOtherMax\%.
The training sample contains almost no joiners, motivating this analysis without establishing that limited exposure caused the remaining errors.

\begin{table*}[t]
\caption{Korean/Chinese extension: false cuts plus missed target cuts per 1,000 Unicode scalars, averaged over \CJKSeeds{} seeds. False cuts include positions inside scalars; the text denominator is shared across encodings. ``Mean'' weights the two languages equally. Both targets and all lookahead conditions are shown. Lower is better.}
\label{tab:cjk}
\centering\small
\begin{tabular}{lrrrrrrr}
\toprule
 &  & \multicolumn{3}{c}{Every cluster} & \multicolumn{3}{c}{Retained parent cuts} \\
\cmidrule(lr){3-5}\cmidrule(lr){6-8}
Encoding & $k$ & Korean & Chinese & Mean & Korean & Chinese & Mean \\
\midrule
UTF-8 & 1 & 0.000 & 0.000 & 0.000 & 54.438 & 159.946 & 107.192 \\
UBE & 1 & 0.035 & 0.046 & 0.041 & \textbf{14.745} & \textbf{103.496} & \textbf{59.121} \\
UBE-BE & 1 & 0.000 & 0.000 & 0.000 & 25.357 & 121.239 & 73.298 \\
UTF-8 & 3 & 0.000 & 0.008 & 0.004 & 11.147 & 103.242 & 57.195 \\
UBE & 2 & 0.096 & 0.023 & 0.060 & 10.556 & 89.276 & 49.916 \\
UBE-BE & 2 & 0.000 & 0.000 & 0.000 & 10.698 & 88.937 & 49.817 \\
\bottomrule
\end{tabular}

\vspace{1.2em}
\caption{Retained parent-tokenizer cuts in eight more scripts, with one predictor per language: false cuts plus missed target cuts per 1,000 Unicode scalars, averaged over \CJKSeeds{} seeds. ``UBE vs.\ UTF-8'' is the relative change at $k=1$. The whole next scalar is visible at $k=3$ for UTF-8 and at $k=2$ for UBE and UBE-BE. Lower is better.}
\label{tab:scriptext}
\begin{tabular}{lrrrrrrr}
\toprule
 & \multicolumn{4}{c}{One symbol ($k=1$)} & \multicolumn{3}{c}{Whole next scalar ($k=3$; $k=2$)} \\
\cmidrule(lr){2-5}\cmidrule(lr){6-8}
Language & UTF-8 & UBE-BE & UBE & UBE vs.\ UTF-8 (\%) & UTF-8 & UBE-BE & UBE \\
\midrule
Japanese & 118.74 & 109.85 & \textbf{38.80} & \textminus{}67.3 & 35.64 & 33.28 & 33.11 \\
Thai & 76.16 & 74.52 & \textbf{19.80} & \textminus{}74.0 & 19.87 & 19.43 & 19.20 \\
Lao & 73.35 & 72.17 & \textbf{6.34} & \textminus{}91.4 & 5.95 & 6.15 & 6.13 \\
Khmer & 53.60 & 54.77 & \textbf{9.31} & \textminus{}82.6 & 5.65 & 5.46 & 5.69 \\
Burmese & 103.06 & 100.17 & \textbf{1.36} & \textminus{}98.7 & 1.18 & 1.30 & 1.33 \\
Hindi & 67.24 & 67.06 & \textbf{0.30} & \textminus{}99.6 & 0.35 & 0.34 & 0.32 \\
Bengali & 80.33 & 77.98 & \textbf{1.33} & \textminus{}98.3 & 1.74 & 2.01 & 1.25 \\
Tamil & 70.24 & 69.34 & \textbf{0.96} & \textminus{}98.6 & 1.16 & 1.22 & 0.91 \\
\bottomrule
\end{tabular}

\end{table*}

\subsection{Korean and Chinese: Different Boundary Targets}
\label{sec:cjk}
Every cluster in the Korean and Chinese NFC evaluation sentences consists of a single scalar; modern Hangul syllables have precomposed NFC forms.
These texts therefore let us separate recovery of grapheme boundaries from selection among them.
In an exploratory extension, we train separate predictors for every cluster boundary, $G(x)$, and for the retained-parent target, $T_{\mathrm{ret}}(x)$, which can merge adjacent clusters into a single unit.
We use Korean and Chinese FineWeb-2 text, \CJKSeeds{} seeds, identical sampled text spans, and matched parameter counts across encodings, excluding training documents that contain an evaluation sentence (Appendix~\ref{app:cjk}).
The extension uses a smaller model and training budget than the preceding probe and is analyzed separately.
Because every cluster here is a single scalar, the every-cluster target reduces to locating scalar ends, which all three encodings expose in the encoded prefix; it does not test the first-symbol question and is reported for completeness.
Balanced error at scalar ends is undefined for that target because there are no negative sites, so we use a common metric for both targets: false cuts plus missed target cuts per 1,000 scalars, including false cuts inside scalars (Table~\ref{tab:cjk}).

UBE shortens the Korean and Chinese input streams by \CJKKoreanSaving\% and \CJKChineseSaving\%, respectively (Appendix Table~\ref{tab:cjkstructure}).
The parent tokenizer splits \CJKKoreanSplit\% and \CJKChineseSplit\% of their clusters; after these internal cuts are deleted, the remaining target still distinguishes which inter-character boundaries to retain.
Stream compression and boundary prediction are therefore evaluated separately.

For the retained-parent target at $k=1$, UBE reduces errors per 1,000 scalars from \CJKKoreanParentUTF{} to \CJKKoreanParentUBE{} in Korean (\CJKKoreanParentReduction\% reduction) and from \CJKChineseParentUTF{} to \CJKChineseParentUBE{} in Chinese (\CJKChineseParentReduction\%).
Every seed favors UBE over both UTF-8 and the same-length UBE-BE control in both languages (Appendix~\ref{app:cjk}).
Relative to UBE-BE, the corresponding reductions are \CJKKoreanParentBEReduction\% and \CJKChineseParentBEReduction\%.
UBE-BE also improves on UTF-8 here, consistent with finer distinctions among Unicode pages: Hangul syllables and the main Han ideograph block span 44 and 82 pages of 256 codepoints, respectively, whereas their UTF-8 representations have only four and six possible first bytes \citep{unicode2025}.
With the whole next scalar visible ($k=2$), UBE and UBE-BE have similar mean errors: their difference changes sign across seeds in each language (Appendix Table~\ref{tab:cjkseeds}) and both sentence-bootstrap intervals span zero (Appendix Table~\ref{tab:cjkcontrasts}), consistent with a byte-order advantage concentrated at one-symbol lookahead.
Comparisons with UTF-8 at $k=3$ still differ in encoded history, sequence length, and optimization, even though all configurations expose the whole next scalar.

A post hoc partition locates the gain at scalar gaps without adjacent whitespace as well as in the aggregate (Appendix~\ref{app:cjkspacing}).
Among scalar gaps with no whitespace on either side, the $k=1$ error rate falls from \CJKKoreanAwayUTF\% to \CJKKoreanAwayUBE\% in Korean and from \CJKChineseAwayUTF\% to \CJKChineseAwayUBE\% in Chinese.
These rates use only the gaps in each stratum, excluding within-scalar positions.
This partition describes where the errors occur; it does not isolate the effect of a language's spacing convention.

\subsection{Eight More Scripts}
\label{sec:scripts}
We repeat the retained-parent comparison in Japanese, Thai, Lao, Khmer, Burmese, Hindi, Bengali, and Tamil, training one predictor per language with the settings, conditions, and \CJKSeeds{} seeds of Section~\ref{sec:cjk} (Table~\ref{tab:scriptext}; Appendix~\ref{app:scriptext}).
In all of these scripts except Japanese, the letters of the main Unicode block share one 256-codepoint page, so the first symbol of UBE-BE, like the first byte of UTF-8, is the same for every letter, whereas UBE's first symbol identifies the letter within the page.
Japanese kana and kanji, like Chinese characters, span many pages.
At one-symbol lookahead, UBE reduces errors relative to UTF-8 in every language and seed; the three-seed mean falls by \XOneMinReduction\% in \XOneMinLanguage{} and up to \XOneMaxReduction\% in \XOneMaxLanguage.
In the single-page scripts, UBE-BE stays within \XOneBEMaxGap\% of UTF-8, so at equal stream length the gain follows byte order; in Japanese, UBE-BE recovers part of the gap (\XOneJapaneseBE{} against \XOneJapaneseUTF{} errors per 1,000 scalars, and \XOneJapaneseUBE{} for UBE).
UBE at $k=1$ already achieves at least \XOneMinGapClosed\% of the error reduction that UTF-8 obtains by extending its lookahead from one byte to the whole next scalar, and the differences between encodings shrink once each sees the whole next scalar.
Retraining the Korean and Chinese retained-parent predictors of Section~\ref{sec:cjk} reproduces their weights exactly and adds evaluations every \XThreeCheckpointEvery{} updates: UBE's seed-mean error at $k=1$ stays below UTF-8 at every checkpoint and reaches UTF-8's final error after \XThreeKoreanCross{} of \CJKSteps{} updates in Korean and \XThreeChineseCross{} in Chinese (Appendix Table~\ref{tab:scriptextcurves}).
In each of the eight new scripts, UBE's seed-mean error at $k=1$ is already below UTF-8's final error at the first checkpoint.

\section{Inherited Segmentation in Released Conversions}
\label{sec:released}
The retained-parent target of Section~\ref{sec:proposal} removes cluster-internal cuts from inherited supervision.
This section measures what released conversions inherit and how to evaluate them: boundary retention (Section~\ref{sec:retained}), sequence lengths under inherited and cluster-aligned targets (Section~\ref{sec:lengths}), sensitivity to span-preserving edits (Section~\ref{sec:edits}), mark--base corruption contrasts (Section~\ref{sec:behaviour}), and the scoring choices that a comparison with the parent requires (Section~\ref{sec:scoring}).

\subsection{Checkpoints and Measurement}
We study three released conversions, two of them paired with their parent models.
Bwen-8B is a conversion of \modelid{Qwen/Qwen3-8B-Base} from the Qwen3 family \citep{qwen3}, although its tokenizer configuration names \modelid{Qwen/Qwen3-8B}; the loaded tokenizers have identical SHA-256 digests in the measurement records.
The second behavioral pair is \modelid{allenai/OLMo-2-0425-1B}, from the OLMo~2 family \citep{olmo2}, and \modelid{allenai/Bolmo-1B}.
We additionally inspect \modelid{allenai/Bolmo-7B}, whose weight lineage starts from Olmo-3-1025-7B \citep{olmo3}.
The latter two conversions share the dolma2 tokenizer but differ in backbone and training history as well as size.
Public revisions and measured runtime identities appear in Appendix~\ref{app:repro}.

We audit natural-text boundaries in float32, processing one unpadded sentence per forward pass through the checkpoint's local encoder and boundary predictor.
The audit reconstructs the exact NFC bytes from parent pieces, validates the cut-offset mapping against shifted alternatives, and verifies the loaded encoder tensors.
The embedding-removal ablation drops the residual parent-vocabulary feature without retraining.
Corruption scores use bfloat16 standalone forwards on a smaller selected item bank; their patch counts are reported separately.

\subsection{Boundary Retention}
\label{sec:retained}
The audit first measures how much inherited segmentation fragments grapheme clusters.
Across the focal languages, Bwen retains at least \MinCutRetention\% of its parent's interior cuts in each language, and its cluster-internal cut count $\ain$ differs from the parent's by at most \MaxInternalChange\%.
This agreement emerges from the predictor's own decisions; the evaluation does not impose parent boundaries.
Retention varies more widely across the control languages (Appendix~\ref{app:boundaries}).

\begin{table}[t]
\caption{Boundary audit totals over all \AuditLanguages{} languages. $N$ counts text units, $A_{\rm in}$ cuts strictly inside a grapheme cluster, and $B_{\rm in}$ cuts strictly inside a Unicode scalar; ``Split'' is the percentage of multi-scalar clusters that receive at least one cut. $^\dagger$Without the parent-vocabulary embedding, no retraining.}
\label{tab:boundary}
\centering\footnotesize
\setlength{\tabcolsep}{3pt}
\begin{tabular}{lrrrr}
\toprule
 & $N$ & $A_{\rm in}$ & $B_{\rm in}$ & Split \% \\
\midrule
Qwen3 parent & 3,370,517 & 2,003,437 & 1,217,605 & 99.89 \\
Bwen-8B & 3,359,694 & 1,995,374 & 1,214,734 & 99.49 \\
Bwen-8B$^\dagger$ & 1,752,896 & 573,822 & 1,578 & 66.18 \\
\bottomrule
\end{tabular}

\end{table}

Across the full audit, Bwen splits \BwenMultiSplit\% of multi-scalar clusters, compared with the parent's \ParentMultiSplit\% (Table~\ref{tab:boundary}).
The close agreement extends to both boundary positions and the proportion of clusters fragmented in this released lineage.

Removing the parent-vocabulary embedding reduces the number of internal cuts in many languages while also coarsening the segmentation: the largest ratio of parent to ablated interior-cut counts is \MaxCoarsening$\times$ (Appendix Table~\ref{tab:ablation}).
The ablated predictor still splits \AblatedMultiSplit\% of the full audit's multi-scalar clusters and makes some within-scalar cuts in Chinese.
Because removal changes both the input features and embedding magnitude without retraining, this intervention demonstrates sensitivity to the vocabulary feature.
It neither establishes a repair at matched granularity nor identifies that feature as the sole source of inherited segmentation.

Patch counts provide complementary evidence on the uncorrupted texts in the mark/base probe.
For Bwen, the pooled patch-to-parent-token ratio is \PatchTokenRatio{}, item-level Pearson correlation is $r=\PatchTokenCorrelation$, and \WithinFivePercent\% of items agree within 5\% relative difference; language-level ratios range over \mbox{\PatchRatioRange}.
On its own bank, Bolmo-1B shows similarly close count agreement with \modelid{OLMo-2-0425-1B}: pooled ratio \PairTwoPatchTokenRatio{}, $r=\PairTwoPatchTokenCorrelation$, and \PairTwoWithinFive\% of items within 5\%, using the parent-tokenizer loading convention in Appendix~\ref{app:loading}.
These smaller-bank measurements record counts only; positional agreement is established by the full Bwen audit.

\subsection{Encoding Length and Patch Density}
\label{sec:lengths}
To assess encoding choices, we distinguish the symbol stream processed by local modules from the patch sequence processed by the global transformer.
Let $P_T(x)$ count the nonempty units defined by a representable interior cut set $T(x)$, and let $L_a(x)$ count symbols under encoding $a$.
The coordinate map in Section~\ref{sec:proposal} preserves the number of cuts, giving
\begin{equation}
  P_T(x)=|T(x)|+1
  \quad\text{independently of }L_a(x).
  \label{eq:accounting}
\end{equation}
Thus a fixed, representable target has the same patch count across encodings, even when its symbol count changes.
The realized cuts of a learned predictor can still depend on the encoding.

\begin{table}[t]
\caption{Positions per NFC codepoint, pooled over the \ProbeLanguages-language panel, with English as a reference. Global computation uses patch positions; local computation uses symbol positions. ``Retained-parent target'' counts the units remaining after deletion of the parent's cluster-internal cuts.}
\label{tab:lengths}
\centering\footnotesize
\setlength{\tabcolsep}{3pt}
\begin{tabular}{lrrr}
\toprule
 & Panel & English & $\times$ English \\
\midrule
\multicolumn{4}{@{}l}{\emph{Global model: patch positions per codepoint}} \\
Parent target & 1.2768 & 0.2093 & 6.10 \\
Bwen-8B, realized & 1.2743 & 0.2057 & 6.19 \\
Retained-parent target & 0.4705 & 0.2093 & 2.25 \\
\multicolumn{4}{@{}l}{\emph{Local modules: byte symbols per codepoint}} \\
UTF-8 bytes & 2.7125 & 1.0010 & 2.71 \\
UBE symbols & 1.8563 & 1.0006 & 1.86 \\
\bottomrule
\end{tabular}

\end{table}

\begin{figure}[t]
\centering
\input{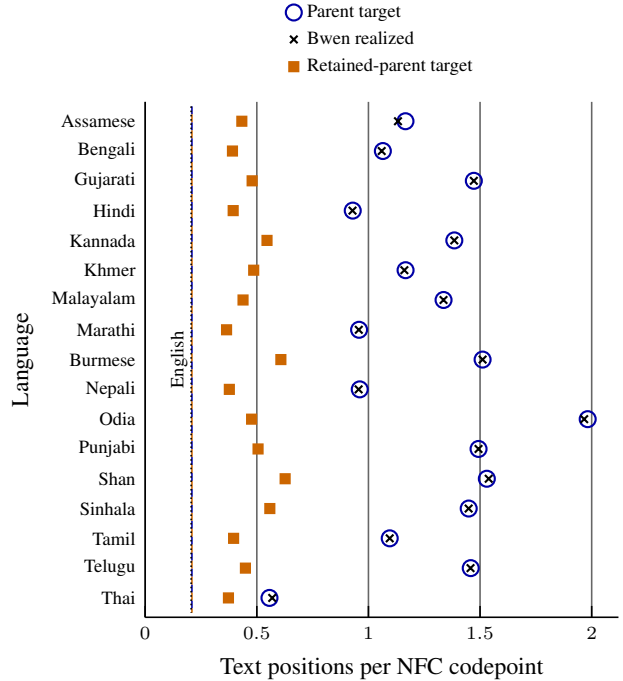}
\pgfmathsetmacro{\DensityTop}{\DensityLastIndex+.5}
\pgfmathsetmacro{\DensityMiddle}{\DensityLastIndex/2}
\begin{tikzpicture}[line width=0.6pt]
\begin{axis}[
  width=0.95\columnwidth,height=8.4cm,line width=0.6pt,
  xmin=0,xmax=2.12,ymin=-.65,ymax=\DensityLastIndex+.65,y dir=reverse,
  xlabel={Text positions per NFC codepoint},ylabel={Language},
  xtick={0,.5,1,1.5,2},
  ytick/.expanded={\DensityTicks},yticklabels/.expanded={\DensityLabels},
  tick label style={font=\scriptsize},label style={font=\small},
  tick style={black,line width=0.6pt},ytick style={draw=none},
  axis lines*=left,axis line style={line width=0.6pt},
  xmajorgrids=true,grid style={black!60,line width=0.6pt},
  every axis plot/.append style={line width=0.6pt},
  legend style={font=\scriptsize,draw=none,at={(.5,1.03)},anchor=south,
                legend columns=1,cells={anchor=west}},
  clip=false]
\addplot[only marks,mark=o,mark size=3pt,blue!65!black,line width=.8pt]
  table[x=parent,y=index]{generated/density.dat};
\addlegendentry{Parent target}
\addplot[only marks,mark=x,mark size=2pt,black,line width=.8pt]
  table[x=bwen,y=index]{generated/density.dat};
\addlegendentry{Bwen realized}
\addplot[only marks,mark=square*,mark size=1.8pt,orange!80!black]
  table[x=snapped,y=index]{generated/density.dat};
\addlegendentry{Retained-parent target}
\draw[blue!65!black,dash pattern=on 3pt off 3pt]
  (axis cs:\ParentEnglish,-.5) -- (axis cs:\ParentEnglish,\DensityTop);
\draw[orange!80!black,dash pattern=on 3pt off 3pt,dash phase=3pt]
  (axis cs:\SnappedParentEnglish,-.5) -- (axis cs:\SnappedParentEnglish,\DensityTop);
\draw[black,densely dotted]
  (axis cs:\BwenEnglish,-.5) -- (axis cs:\BwenEnglish,\DensityTop);
\node[font=\scriptsize,rotate=90,anchor=south,fill=white,inner sep=1pt]
  at (axis cs:\BwenEnglish,\DensityMiddle) {English};
\end{axis}
\end{tikzpicture}
\caption{Text positions per NFC codepoint on full devtest, excluding BOS. Parent-target and realized Bwen markers nearly coincide. The retained-parent target deletes cluster-internal parent cuts. Vertical English references are nearly coincident; the two target densities are identical in English.}
\label{fig:density}
\end{figure}

\paragraph{Realized versus target density.}
On full devtest, Bwen's own predictor produces \BwenDensity{} text units per NFC codepoint on the \ProbeLanguages-language panel and \BwenEnglish{} in English, a ratio of \BwenRatio{} (Table~\ref{tab:lengths}).
The corresponding ratio for the parent's segmentation is \ParentRatio{}.
Figure~\ref{fig:density} separates realized Bwen density from the parent and retained-parent target densities in each language.
All three densities count interior cuts plus one per sentence, excluding BOS.
These densities measure representation length; they do not imply runtime ratios or equal semantic content per codepoint.

\paragraph{Stream length and vocabulary size.}
UBE shortens the panel's symbol stream by \UBESaving\%.
Using UBE in Bolmo's fused symbol/flag output head would require $(4+512)\times2=1{,}032$ identifiers.
At the recorded hidden width, expanding the untied input and output tables adds \AlphabetMillions{} million parameters, or \AlphabetShare\% of Bwen's measured total (Appendix~\ref{app:cost}).
For comparison, its existing parent-vocabulary embedding contains \EmbeddingMillions{} million parameters, \EmbeddingShare\% of the model.
The wider head changes both parameter count and per-symbol work.

\paragraph{Compatibility with inherited boundaries.}
Odia has \OdiaTokens{} parent tokens but only \OdiaSymbols{} UBE symbols.
Each nonempty patch must contain at least one symbol, so this stream cannot preserve every parent token as a separate patch.
Shorter encodings also remove some within-scalar positions at which the parent tokenizer can cut.
Having enough symbols in aggregate would therefore be necessary, but not sufficient, to preserve the original cut offsets.

Independent decoding of parent pieces also fails to provide a general mapping: \InvalidPieces{} of the parent's \VocabularyPieces{} ordinary vocabulary pieces are not valid standalone UTF-8 strings; byte-level tokenizers place no UTF-8 constraint on token formation \citep{firestone2025illformed}.
Their embedding identifiers remain usable, but transferring those features to new symbol positions requires an explicit alignment policy.
The cluster-aligned construction in Section~\ref{sec:proposal} resolves the representability of the boundary target; it leaves vocabulary-feature transfer as a separate design choice.

\subsection{Sensitivity to Span-Preserving Edits}
\label{sec:edits}
Natural-text boundary agreement does not determine how segmentation responds to edits.
We construct paired mark and base substitutions within the same multi-scalar cluster, each replacing one scalar with another of the same script association and UTF-8 width while preserving NFC and every grapheme span.
At least one substitution changes Bwen's patch count in \UnstableItems{} of \ProbeItems{} items (\InstabilityFull\%).
A count change establishes a segmentation change, although an unchanged count can conceal moved boundaries.

\begin{table}[t]
\caption{Instability in patch counts under single-scalar, span-preserving edits, measured with standalone forwards on \SubsetPerLanguage{} items per language. Flag share is measured on the original text. Bytes/patch uses all sentence bytes and counts the forced BOS flag in the denominator. The dolma2 conversions share an item bank; Bwen uses different substitutions and prefixes. Bwen's row uses the first \SubsetPerLanguage{} of its \ProbeItemsPerLanguage{} items per language (all \ProbeItems{} items: \InstabilityFull\%, Section~\ref{sec:edits}); flag share pools path cost over items.}
\label{tab:instability}
\centering\footnotesize
\setlength{\tabcolsep}{3pt}
\begin{tabular}{lrrrr}
\toprule
Model & Items changed & \% & Flag share (\%) & Bytes/patch \\
\midrule
Bwen-8B & 458/850 & 53.9 & 13.19 & 2.127 \\
Bolmo-1B & 328/850 & 38.6 & 9.76 & 1.746 \\
Bolmo-7B & 326/850 & 38.4 & 10.46 & 1.746 \\
\bottomrule
\end{tabular}

\end{table}

Bolmo-1B and Bolmo-7B show instability rates of \InstabilityOneB\% and \InstabilitySevenB\% on their matched bank (Table~\ref{tab:instability}).
Their instability indicators agree on \CheckpointAgreement{} of \MatchedItems{} items despite a \CheckpointSizeRatio$\times$ parameter difference.
The pattern recurs across separately trained checkpoints from the same group and conversion recipe, with the same parent tokenizer.
Because text, targets, and models are not varied independently, this agreement does not identify which factor determines the response.

Across language aggregates, bytes per patch correlates positively with instability in each checkpoint ($r$ from \BppAssociationRange{}).
Short average patches do not ensure stability: Bolmo-1B has \ShanBytesPerPatch{} bytes per patch in Shan with \ShanInstability\% instability, and \OdiaBytesPerPatch{} in Odia with \OdiaInstability\% instability.
Equal byte lengths would force equal patch counts if every byte formed a patch in both strings; a short average patch length on the originals is a weaker condition.

The dolma2 pair's language-level flag shares also correlate closely ($r=\FlagCorrelation$), but flag-share/instability associations are near zero for Bwen and strongly positive for that pair (Appendix~\ref{app:instability}, Table~\ref{tab:associations}).

\subsection{Mark--Base Corruption Contrasts}
\label{sec:behaviour}
Using the same span-preserving edits, we examine whether conversion changes the relative penalty for corrupting a mark versus a base.
The probe replaces a mark (Unicode category \texttt{Mn} or \texttt{Mc}) or a base letter (\texttt{Lo}) within the same cluster; both conditions therefore alter cluster content.
Eligible replacements are distinct scalars of the same UTF-8 width drawn from the language's script-associated inventory, and must preserve the first parent token and all grapheme spans.
The selected \SourceSentences{} parallel source sentences yield \ProbeItems{} items per parent/conversion pair.
Each pair shares an item bank, while the banks differ between pairs.

For the representation-path cost $\ell$, define margins $m=\ell(x_{\rm mark})-\ell(x)$ and $b=\ell(x_{\rm base})-\ell(x)$.
The primary statistic, $R=\overline m/\overline b$, compares mean mark and base margins.
The secondary statistic, $I=\operatorname{mean}[(b-m)/(|b|+|m|)]$, averages the normalized contrast over items for which it is defined.
All reported mean base margins are positive, so larger $R$ indicates a larger mean mark penalty relative to the mean base penalty; $R$ can exceed one.
Smaller $I$ indicates greater relative mark penalties under item-level normalization.
The summaries weight items differently and can move in different directions.
Substitution frequency, orthographic legality, and contextual predictability also affect these penalties, limiting their interpretation as measures of character processing.

\begin{table}[t]
\caption{Pooled margin-ratio differences, conversion minus parent. Intervals are paired percentile bootstrap intervals over shared FLORES source sentences, pointwise and without multiplicity correction. ``Excl.\ 0'' counts languages whose own interval excludes zero.}
\label{tab:probe}
\centering\footnotesize\setlength{\tabcolsep}{2.5pt}
\begin{tabular}{lrcc}
\toprule
Pair & $\Delta R$ & 95\% interval & Excl.\ 0 \\
\midrule
Qwen3 $\to$ Bwen-8B & +0.0177 & [+0.0074, +0.0283] & 7/17 \\
OLMo-2 $\to$ Bolmo-1B & +0.0064 & [\textminus{}0.0080, +0.0214] & 2/17 \\
\bottomrule
\end{tabular}

\end{table}

The Qwen/Bwen difference is \PairOneDelta{} with 95\% interval \PairOneCI{} (Table~\ref{tab:probe}); its item-contrast difference is \PairOneItemDelta{} \PairOneItemCI{}.
For OLMo/Bolmo, the estimates are \PairTwoDelta{} \PairTwoCI{} and \PairTwoItemDelta{} \PairTwoItemCI{}, respectively.
Both second-pair intervals contain zero, and the ratio interval also contains the first pair's point estimate.
Separate tests against zero neither establish equivalence nor test whether the two pairs differ.

Language-level ratio differences agree in sign in \SignAgreement{} of \ProbeLanguages{} languages (Appendix~\ref{app:probe}).
Their Pearson correlation is \CrossPairAll{}, changing to \CrossPairOmitMyanmar{} when the two Myanmar-script languages are omitted.
The language estimates are noisy and dependent, precluding an interpretation as independent draws for a sign test.
Within pair one, omitting Thai changes the pooled point estimate to \OmitThaiDelta{} (Appendix Table~\ref{tab:omissions}).
Thai's control designation comes from the audit and does not imply a null effect in this probe.
The primary analysis retains all eligible languages; omissions describe sensitivity of the pooled estimate.

These estimates do not establish a consistent shift in relative mark penalties across the two parent/conversion pairs, but remain compatible with effects the probe does not resolve.
They combine changes in segmentation and scoring because the edits can change the boundary path.

\subsection{Scoring Parent Models and Byte Conversions}
\label{sec:scoring}
Interpreting the corruption contrasts requires three scoring choices to be explicit (Appendix~\ref{app:pitfalls}).
First, the scored byte suffix must match the parent's scored tokens under an explicit prefix convention: prepending the parent's declared BOS raises the cost of otherwise unchanged targets by \BOSInflation\%.
Second, batch composition can alter patch positions and decisions in the released wrappers; it changes Bwen's selected boundary path in \BatchPathChanges{} of \BatchVariants{} tested variants.
We therefore score each variant in a standalone forward pass.
Third, the joint byte/flag score conditions on one selected boundary history.
It measures the cost of that path, whereas text likelihood requires marginalization over alternative boundary histories.
The same distinction holds for tokenized models: a string's probability sums over every token sequence that decodes to it, not only the canonical one \citep{cao2021marginal,chirkova2023marginalize,vieira2025canonical}.
Our parent score likewise follows one canonical tokenization (Appendix~\ref{app:scoring}).

\section{Related Work}
\paragraph{Byte and hierarchical models.}
Models operating below the subword level differ in both representation and allocation of computation.
ByT5 uses bytes in a text-to-text transformer \citep{byt5}, CANINE downsamples character inputs \citep{canine}, and MambaByte uses a selective state-space model \citep{mambabyte}.
MEGABYTE separates local and global computation with fixed patches \citep{megabyte}; SpaceByte places global computation at ``spacelike'' positions such as UTF-8 leading bytes, whitespace, and punctuation \citep{spacebyte}; BLT selects patches using entropy \citep{blt}.
Learned segmentation takes several forms: dynamic token pooling learns boundaries or supervises them with subword or entropy targets \citep{dynamicpooling}, MrT5 learns deletion \citep{mrt5}, and H-Net learns chunking \citep{hnet}; AU-Net instead builds an autoregressive hierarchy over rule-based word splits \citep{aunet}.
MAGNET routes each byte sequence through a boundary predictor for its script, with per-script compression targets chosen to make segmentation granularity similar across scripts \citep{ahia2024magnet}.
FLEXITOKENS allows a range of boundary rates during adaptation \citep{flexitokens}, while Fast BLT and dynamic multi-byte prediction target decoding efficiency \citep{fastblt,mbp}.
These architectures treat compression and boundary placement as explicit design choices.
Our analysis concerns the information that an encoding exposes to the boundary predictor, particularly when supervision is inherited from a subword model.
The distinctions we study, such as a dependent mark versus a letter after the same base, arise within one script and usually within one 256-codepoint page, so routing by script, as in MAGNET, does not supply them.
The study of entropy-based patch allocation at matched patch counts by \citet{boundarydependence} addresses a complementary question, and SOMBRERO scores boundaries by how strongly chunk starts concentrate on bytes that are hard to predict \citep{sombrero2026}; our audit uses grapheme integrity as the boundary criterion.

\paragraph{Transfer and encoding.}
Bolmo transfers a pretrained model into a byte architecture through distillation \citep{bolmo}, whereas zero-shot tokenizer transfer predicts embeddings for a replacement tokenizer \citep{zett}.
H-Net similarly distills Llama~3.2 3B into a byte-level model around its frozen backbone, supervising the routing module with the parent's token starts and noting that these boundaries cannot always be predicted online \citep[Appendix~E.2]{hnet}.
Inference-time methods instead expose a tokenized model as a byte- or character-level distribution without training \citep{phan2025exact,vieira2025characters,hayase2026bytesampler}; byteification trains local modules and a boundary predictor around the parent transformer.
Concurrent work adapts H-Net-style dynamic byte chunking to a frozen subword model, pulling chunk embeddings toward averaged subword embeddings and adding part-of-speech tagging supervision to guide word-level boundaries \citep{tokenizersfail2026}.
Vocabulary expansion based on hidden-state similarity can consolidate multi-token characters in a frozen model \citep{tokenpremiums}.
Closer to our boundary question, \citet{haltiuk} reports nearly inherited Bolmo segmentation, proposes capability-preserving boundary transfer, and asks which shared alphabet to use.
We give cluster-aligned target constructions and analyze their predictability at limited lookahead.

Alternative encodings provide another way to change the representation presented to a tokenizer or model.
MYTE uses morpheme inventories \citep{myte}, and SCRIPT combines Unicode script/category structure with constrained merges \citep{script}.
BBPE16 motivates UTF-16LE byte tokenization by removing the format bits of three-byte UTF-8 encodings and evaluates BPE on those bytes in multilingual speech recognition \citep{bbpe16}.
Bit-level BPE repacks and deduplicates UTF-8 prefix bits in Korean, Chinese, and Japanese byte fallbacks \citep{bitbpe}; Latom learns corpus-dependent atomic character codes \citep{latom}.
UBE routes scalars through UTF-8 or UTF-16LE according to their UTF-8 length, assigning the routes disjoint alphabets \citep{ube}.
We compare this representation with UTF-8 and a matched-length big-endian control to study boundary prediction across scripts at limited lookahead.

\paragraph{Script-aware segmentation.}
Grapheme pair encoding initializes BPE with graphemes \citep{gpe}, MorphTok constrains dependent-vowel segmentation \citep{morphtok}, type-driven tokenization extends tokens to orthographic boundaries \citep{typedriven}, and WWHO preserves rule-defined syllables \citep{wwho}.
These constraints refer to different written units.
Indic grapheme parsers model orthographic syllables \citep{indicparser}, while a Devanagari grapheme-cluster tokenizer has been evaluated on Hindi and Marathi tasks \citep{graphemetokenizer}.
Word-pooled tokenization groups bytes or characters at word boundaries \citep{wordpooled}; HAT uses whitespace or Unicode word splits \citep{hat}.
Korean methods exploit Jamo in character representations or tokenization \citep{kombo,jamobpe}, or use morpheme analysis to constrain subword segmentation \citep{morphemematters}.
Our target construction uses Unicode grapheme clusters to constrain inherited cuts, while the predictor analysis asks whether the available encoded prefix distinguishes the required decisions.

\paragraph{Korean and Chinese fragmentation.}
H-Net evaluates learned chunking against a SpaceByte-style rule on Chinese text without regular word spaces \citep{hnet}.
Our whitespace partition instead characterizes errors in predicting retained parent cuts; its targets are inherited subwords, not linguistic word labels.
Thunder-Tok filters vocabulary candidates that mix complete characters with incomplete byte fragments but allows isolated fragments of a single character \citep{thundertok}, so it does not guarantee grapheme integrity.
Constructed combinations of incomplete tokens can cause failures \citep{improbablebigrams}, yet single-token Chinese characters can also perform worse on character-structure tasks \citep{haslettchinese}.
These task-dependent findings support evaluating boundary prediction, symbol compression, and linguistic quality as distinct outcomes; morphological alignment, measured across 70 languages, likewise explains little of the variance in the downstream performance of five pretrained models \citep{arnett2025morphscore}.

\paragraph{Disparity and character probes.}
Controlled tokenizer comparisons, multilingual token lengths, and commercial token pricing document disparities in representation cost \citep{rust,petrov,ahia}.
The Southeast Asian comparison by \citet{lee} also finds unequal language-level patch parity for BLT, extending this concern to entropy-based patching.
Analysis of UTF-8's efficiency and equity tradeoffs motivates considering alternative input encodings \citep{goot}.
Work on vowel signs identifies letter-category pretokenization as one source of fragmentation in abugidas \citep{vowels}, supporting the distinction between scalar and cluster boundaries without explaining every observed cut.

Character-level evaluation provides a separate perspective on these representations.
CUTE tests orthographic knowledge and manipulation \citep{cute}.
Bolmo reports approximately \BolmoCuteMillions{} million tokens of English-only CUTE-style tasks (about \BolmoCutePercent\% of its data), augmenting approximately \BolmoMixBillions{} billion Dolma~3 tokens counted with the Dolma2 tokenizer \citep[Section~4 and Appendix~C]{bolmo}.
Our mark/base substitutions measure relative corruption penalties, which complement task-based character evaluation but do not provide a general character-understanding score.

\section{Limitations}
The lookahead evidence comes from small causal predictors with \AOneSeeds{} seeds on the \AOneLanguages-language panel and \CJKSeeds{} on the Korean/Chinese extension; the sentence-bootstrap intervals do not capture variation over training runs.
Text exposure is matched across encodings but compute is not, and UBE-BE changes the entire encoded history as well as the first lookahead symbol.
The first-symbol argument covers continuations within a known Unicode page; cross-page collisions preclude a universal one-symbol guarantee.
The released-model analyses cover two parent families and one family of conversion recipes on FLORES-200 devtest translations, and because the checkpoints differ in architecture, size, data, and training history, their comparison does not isolate the effect of byte-level modeling.

\section{Conclusion}
Boundary prediction depends on how an encoding exposes the distinctions relevant to a writing system and segmentation target.
Bwen largely reproduces its parent's cuts, including those inside grapheme clusters, and Bolmo-1B's patch counts track its parent's token counts; restricting inherited cuts to grapheme boundaries removes this fragmentation from supervision and yields targets representable across encodings.
For scalars requiring three UTF-8 bytes, UBE's UTF-16 route removes the format-bit overhead and places the low byte first, making within-page character distinctions available earlier.
At one-symbol lookahead, the trained UBE and UBE-BE predictors support the importance of byte order for grapheme boundary recovery and for retained subword boundary selection.
In Korean and Chinese, where the every-cluster target is already recoverable from scalar ends, UBE reduces retained-parent errors relative to UTF-8 by \CJKKoreanParentReduction\% and \CJKChineseParentReduction\% at the same lookahead; in eight more scripts, by \XOneMinReduction\% to \XOneMaxReduction\%.
Which byte comes first, not only how many bytes a scalar takes, is therefore a design choice for byte-level boundary predictors.

\FloatBarrier
\label{main-end}
\ifpreprint
\section*{Use of Generative AI}
Generative AI tools helped draft and revise the text and code; the author verified all content and takes full responsibility for it.
\fi
\section*{Impact Statement}
\label{impact-start}
This work examines how text encodings affect boundary prediction and representation length across writing systems.
It provides ways to identify and measure design choices that fragment written units or disproportionately lengthen some languages' representations.
The selected languages and translated corpus do not cover all users, dialects, or writing practices, and Unicode grapheme clusters provide an operational definition of written units with language-specific limits.
Changes in representation can redistribute computational cost and model quality across scripts.
Assessing their social benefit requires native-language evaluation that reports both gains and regressions.

\label{references-start}
\bibliography{references}
\bibliographystyle{icml2026}
\clearpage
\appendix
\onecolumn
\raggedbottom
\widowpenalty=10000
\section{Full Boundary Audit}
\label{app:boundaries}
Tables~\ref{tab:boundaryfull} and \ref{tab:ablation} report the complete \AuditLanguages-language audit.
An asterisk marks the languages designated as controls in the audit.
The ranking that selects the focal languages uses the concatenated NFC devtest sentences of each \mbox{FLORES-200} language with at least 200 non-ASCII scalars; a scalar counts when a Qwen3 token boundary falls strictly inside its UTF-8 bytes.
Offsets refer to the NFC text reconstructed from the parent tokenizer's byte pieces.
$K$ counts grapheme clusters and $N$ counts text units; $S$ abbreviates $\ain-\bin$.
``Shared'' is the fraction of the parent's interior cuts also predicted by Bwen, with denominator $N_P$ minus the number of sentences.
It measures directional retention rather than symmetric boundary similarity.

\begin{table}[htbp]
\caption{Parent-tokenizer and predicted Bwen boundaries on full devtest. Grapheme clusters are the units of the fragmentation audit; they need not coincide with linguistic syllables.}
\label{tab:boundaryfull}
\centering\footnotesize\setlength{\tabcolsep}{2.4pt}
\begin{tabular}{lrrrrrrrrrr}
\toprule
 &  & \multicolumn{4}{c}{Parent (Qwen3)} & \multicolumn{4}{c}{Bwen-8B} &  \\
\cmidrule(lr){3-6}\cmidrule(lr){7-10}
Language & $K$ & $N$ & $A_{\rm in}$ & $B_{\rm in}$ & $S$ & $N$ & $A_{\rm in}$ & $B_{\rm in}$ & $S$ & Shared \% \\
\midrule
Amharic & 87,132 & 111,737 & 40,880 & 40,880 & 0 & 111,600 & 40,752 & 40,752 & 0 & 99.87 \\
Assamese & 83,536 & 148,763 & 93,534 & 54,661 & 38,873 & 144,474 & 92,098 & 54,662 & 37,436 & 97.07 \\
Bengali & 81,693 & 138,814 & 87,846 & 46,927 & 40,919 & 138,133 & 87,858 & 46,948 & 40,910 & 99.47 \\
English$^*$ & 131,966 & 27,621 & 0 & 0 & 0 & 27,149 & 0 & 0 & 0 & 97.71 \\
Gujarati & 80,578 & 186,119 & 125,560 & 79,712 & 45,848 & 186,117 & 125,559 & 79,711 & 45,848 & 99.99 \\
Hindi & 85,957 & 121,957 & 70,261 & 31,358 & 38,903 & 121,821 & 70,299 & 31,393 & 38,906 & 99.83 \\
Kannada & 90,371 & 191,127 & 115,813 & 68,157 & 47,656 & 191,102 & 115,813 & 68,157 & 47,656 & 99.97 \\
Khmer & 89,549 & 178,695 & 104,228 & 51,767 & 52,461 & 177,865 & 103,734 & 51,955 & 51,779 & 99.19 \\
Malayalam & 79,428 & 199,583 & 134,139 & 80,601 & 53,538 & 199,552 & 134,106 & 80,568 & 53,538 & 99.98 \\
Marathi & 80,288 & 127,476 & 78,983 & 36,756 & 42,227 & 127,298 & 78,984 & 36,758 & 42,226 & 99.84 \\
Burmese & 101,038 & 246,611 & 147,448 & 85,317 & 62,131 & 246,611 & 147,282 & 85,151 & 62,131 & 99.87 \\
Nepali & 78,723 & 121,958 & 74,132 & 35,053 & 39,079 & 121,326 & 74,132 & 35,054 & 39,078 & 99.47 \\
Odia & 82,869 & 267,893 & 203,475 & 151,194 & 52,281 & 265,768 & 201,453 & 149,172 & 52,281 & 99.20 \\
Punjabi & 92,928 & 200,987 & 132,946 & 91,281 & 41,665 & 200,965 & 132,942 & 91,277 & 41,665 & 99.93 \\
Russian$^*$ & 142,095 & 48,471 & 45 & 44 & 1 & 44,015 & 40 & 39 & 1 & 83.10 \\
Shan & 129,732 & 288,136 & 170,112 & 111,572 & 58,540 & 289,716 & 169,485 & 110,934 & 58,551 & 99.59 \\
Sinhala & 92,829 & 189,956 & 116,779 & 78,528 & 38,251 & 189,913 & 116,751 & 78,500 & 38,251 & 99.97 \\
Tamil & 99,724 & 168,828 & 107,794 & 53,387 & 54,407 & 168,814 & 107,794 & 53,387 & 54,407 & 99.99 \\
Telugu & 76,567 & 193,169 & 133,702 & 77,809 & 55,893 & 193,046 & 133,699 & 77,806 & 55,893 & 99.92 \\
Thai$^*$ & 99,271 & 70,202 & 23,179 & 22 & 23,157 & 71,752 & 20,103 & 22 & 20,081 & 88.67 \\
Tigrinya & 90,698 & 114,713 & 42,271 & 42,269 & 2 & 114,618 & 42,183 & 42,181 & 2 & 99.91 \\
Chinese (Simpl.)$^*$ & 43,248 & 27,701 & 310 & 310 & 0 & 28,039 & 307 & 307 & 0 & 91.81 \\
\bottomrule
\end{tabular}

\end{table}

\begin{table}[htbp]
\caption{Parent-vocabulary embedding removal without retraining. Coarsening is the ratio of parent to ablated interior-cut counts; values below one indicate more cuts in the ablated predictor than in the parent segmentation. $\ain$/cut is the fraction of interior cuts that fall strictly inside a cluster. A dash marks an undefined split percentage because no multi-scalar clusters occur.}
\label{tab:ablation}
\centering\small\setlength{\tabcolsep}{4pt}
\begin{tabular}{lrrrrrrr}
\toprule
Language & $N$ & $A_{\rm in}$ & $B_{\rm in}$ & $S$ & $A_{\rm in}$/cut & Coarsening & Multi split \% \\
\midrule
Amharic & 34,434 & 0 & 0 & 0 & 0.0000 & 3.31 & --- \\
Assamese & 103,183 & 37,498 & 0 & 37,498 & 0.3670 & 1.45 & 88.07 \\
Bengali & 107,247 & 41,026 & 0 & 41,026 & 0.3862 & 1.30 & 88.31 \\
English$^*$ & 49,290 & 0 & 0 & 0 & 0.0000 & 0.55 & --- \\
Gujarati & 103,873 & 45,243 & 0 & 45,243 & 0.4398 & 1.80 & 98.61 \\
Hindi & 108,523 & 44,480 & 0 & 44,480 & 0.4137 & 1.12 & 98.53 \\
Kannada & 119,574 & 46,383 & 0 & 46,383 & 0.3912 & 1.60 & 97.34 \\
Khmer & 18,608 & 9 & 0 & 9 & 0.0005 & 10.10 & 0.02 \\
Malayalam & 131,108 & 68,020 & 0 & 68,020 & 0.5228 & 1.53 & 97.00 \\
Marathi & 116,563 & 52,068 & 0 & 52,068 & 0.4506 & 1.09 & 98.64 \\
Burmese & 20,373 & 23 & 0 & 23 & 0.0012 & 12.69 & 0.05 \\
Nepali & 110,886 & 47,631 & 0 & 47,631 & 0.4335 & 1.10 & 98.82 \\
Odia & 87,027 & 26,495 & 0 & 26,495 & 0.3080 & 3.10 & 54.31 \\
Punjabi & 108,367 & 41,081 & 0 & 41,081 & 0.3827 & 1.86 & 98.55 \\
Russian$^*$ & 47,713 & 1 & 0 & 1 & 0.0000 & 1.02 & 100.00 \\
Shan & 19,395 & 32 & 0 & 32 & 0.0017 & 15.62 & 0.06 \\
Sinhala & 77,275 & 15,059 & 0 & 15,059 & 0.1975 & 2.48 & 39.43 \\
Tamil & 91,376 & 26,821 & 0 & 26,821 & 0.2968 & 1.86 & 49.30 \\
Telugu & 112,569 & 54,919 & 0 & 54,919 & 0.4923 & 1.72 & 98.42 \\
Thai$^*$ & 112,515 & 25,455 & 0 & 25,455 & 0.2283 & 0.62 & 96.13 \\
Tigrinya & 38,387 & 0 & 0 & 0 & 0.0000 & 3.04 & 0.00 \\
Chinese (Simpl.)$^*$ & 34,610 & 1,578 & 1,578 & 0 & 0.0470 & 0.79 & --- \\
\bottomrule
\end{tabular}

\end{table}

Removing the vocabulary feature changes both the input distribution and embedding magnitude while leaving the byte embeddings and encoder fixed (Section~\ref{sec:retained}).
\FloatBarrier

\section{Full Sequence-Length Accounting}
\label{app:cost}
Table~\ref{tab:costfull} covers the \ProbeLanguages-language panel and three reference languages.
$M$ counts NFC scalars, including whitespace and punctuation.
Parent-unit counts and scalar denominators agree exactly with the boundary audit for every shared language.
The ``Retained'' column reports the retained-parent target density, $(N_P-\ain(P))/M$, computed from the parent cuts and grapheme boundaries.
It describes the target segmentation rather than a learned predictor's realized cuts.

\begin{table}[htbp]
\caption{Sequence-length accounting for the full panel and reference languages. Scalars $M$ are NFC codepoints, and Parent $N$ counts parent tokens. Symbol columns give UTF-8 byte and UBE symbol counts. Positions per codepoint distinguish parent-target density, realized Bwen density, and retained-parent target density (``Retained''). UBE saving is the percentage reduction in symbol count.}
\label{tab:costfull}
\centering\small\setlength{\tabcolsep}{3.2pt}
\begin{tabular}{lrrrrrrrr}
\toprule
 &  &  & \multicolumn{2}{c}{Symbols} & \multicolumn{3}{c}{Positions per codepoint} &  \\
\cmidrule(lr){4-5}\cmidrule(lr){6-8}
Language & Scalars $M$ & Parent $N$ & UTF-8 & UBE & Parent & Bwen & Retained & UBE saving \% \\
\midrule
Assamese & 127,594 & 148,763 & 340,036 & 233,823 & 1.1659 & 1.1323 & 0.4328 & 31.2 \\
Bengali & 130,391 & 138,814 & 348,729 & 239,566 & 1.0646 & 1.0594 & 0.3909 & 31.3 \\
Gujarati & 126,428 & 186,119 & 331,068 & 228,751 & 1.4721 & 1.4721 & 0.4790 & 30.9 \\
Hindi & 131,180 & 121,957 & 337,439 & 234,314 & 0.9297 & 0.9287 & 0.3941 & 30.6 \\
Kannada & 138,027 & 191,127 & 375,341 & 256,686 & 1.3847 & 1.3845 & 0.5456 & 31.6 \\
Khmer & 153,212 & 178,695 & 435,160 & 294,202 & 1.1663 & 1.1609 & 0.4860 & 32.4 \\
Malayalam & 149,332 & 199,583 & 411,741 & 280,538 & 1.3365 & 1.3363 & 0.4382 & 31.9 \\
Marathi & 133,051 & 127,476 & 355,511 & 244,286 & 0.9581 & 0.9568 & 0.3645 & 31.3 \\
Burmese & 163,169 & 246,611 & 465,528 & 314,355 & 1.5114 & 1.5114 & 0.6077 & 32.5 \\
Nepali & 126,903 & 121,958 & 339,472 & 233,192 & 0.9610 & 0.9561 & 0.3769 & 31.3 \\
Odia & 135,150 & 267,893 & 362,392 & 248,774 & 1.9822 & 1.9665 & 0.4766 & 31.4 \\
Punjabi & 134,593 & 200,987 & 347,676 & 241,139 & 1.4933 & 1.4931 & 0.5055 & 30.6 \\
Shan & 188,284 & 288,136 & 522,478 & 355,396 & 1.5303 & 1.5387 & 0.6268 & 32.0 \\
Sinhala & 131,080 & 189,956 & 346,056 & 238,571 & 1.4492 & 1.4488 & 0.5583 & 31.1 \\
Tamil & 154,131 & 168,828 & 421,635 & 287,885 & 1.0954 & 1.0953 & 0.3960 & 31.7 \\
Telugu & 132,465 & 193,169 & 353,661 & 243,067 & 1.4583 & 1.4573 & 0.4489 & 31.3 \\
Thai & 126,109 & 70,202 & 364,857 & 245,493 & 0.5567 & 0.5690 & 0.3729 & 32.7 \\
English & 131,966 & 27,621 & 132,096 & 132,047 & 0.2093 & 0.2057 & 0.2093 & 0.0 \\
Russian & 142,096 & 48,471 & 260,342 & 260,231 & 0.3411 & 0.3098 & 0.3408 & 0.0 \\
Chinese (Simpl.) & 43,248 & 27,701 & 120,803 & 82,081 & 0.6405 & 0.6483 & 0.6333 & 32.1 \\
\bottomrule
\end{tabular}

\end{table}

\paragraph{Parameter arithmetic.}
The recorded Bwen configuration has hidden width \HiddenWidth{}, an output vocabulary of 520 identifiers, and \ParentRows{} parent-embedding rows.
The byte input embedding and untied output head each have shape $520\times\text{\HiddenWidth}$.
The parent-vocabulary table has shape $\text{\ParentRows}\times\text{\HiddenWidth}$, and the measured complete model contains \BwenParameters{} parameters.
Extending both alphabet-facing tables to 1,032 rows adds $2(1{,}032-520)\times\text{\HiddenWidth}$ parameters, or \AlphabetMillions{} million (\AlphabetShare\%).
The parent-vocabulary embedding accounts for \EmbeddingMillions{} million (\EmbeddingShare\%).
These parameter counts exclude optimizer state and activation memory and do not quantify per-symbol computation.

\paragraph{Different patch-count conventions.}
The float32 audit counts interior cuts plus one final text unit and excludes the separate BOS patch.
The bfloat16 patch-count diagnostic instead sums boundary flags, including the forced BOS flag but suppressing the final text byte's flag because lookahead is unavailable.
These conventions give the same numerical count when interior cuts agree: each adds one to the number of interior cuts.
The diagnostic's comparison with ordinary parent-token counts therefore requires no additional BOS adjustment.
It uses the recorded sums on its selected item bank, distinct from the full-devtest densities normalized to English.
\FloatBarrier

\section{Lookahead Encoding Examples}
\label{app:lookahead}
Table~\ref{tab:lookahead} gives concrete continuation pairs for P5.
For each row, use the listed consonant as the preceding base.
Appending the vowel sign extends that base's cluster, whereas appending another copy of the consonant starts a new cluster.
The continuations therefore require opposite decisions after identical left context.
They share the first UTF-8 byte but have distinct first UTF-16LE bytes within their common 256-codepoint page.

\begin{table}[htbp]
\caption{Hexadecimal continuation scalars and their first byte values under UTF-8 and UTF-16LE. The UTF-16LE values occupy UBE's separate UTF-16 symbol alphabet.}
\label{tab:lookahead}
\centering\small
\begin{tabular}{lcccc}
\toprule
Script & Vowel sign & Consonant & UTF-8 first (sign / base) & UTF-16LE first (sign / base) \\
\midrule
Devanagari & U+093F & U+092E & E0 / E0 & 3F / 2E \\
Bengali & U+09BF & U+09AE & E0 / E0 & BF / AE \\
Oriya & U+0B3F & U+0B2E & E0 / E0 & 3F / 2E \\
Sinhala & U+0DD2 & U+0DB8 & E0 / E0 & D2 / B8 \\
Thai & U+0E34 & U+0E21 & E0 / E0 & 34 / 21 \\
Myanmar & U+102D & U+1019 & E1 / E1 & 2D / 19 \\
Khmer & U+17B7 & U+1798 & E1 / E1 & B7 / 98 \\
\bottomrule
\end{tabular}

\end{table}

Every scalar from U+0800 through U+0FFF begins with UTF-8 \texttt{E0}; U+1000 through U+1FFF begins with \texttt{E1}.
U+200D and U+090D lie on different pages but share the UTF-16LE low byte \texttt{0D}: after an ordinary Devanagari base, the first extends the cluster and the second starts a new one, a remaining ambiguity under UBE.
The native UBE encoder routes all listed scalars through UTF-16LE, emitting the low byte first.
Together with the causal encoder and suffix lookup assumed in P5, the shared first symbol establishes indistinguishability for the corresponding every-cluster decisions.

\begin{figure}[ht]
\centering
\includegraphics[width=\textwidth]{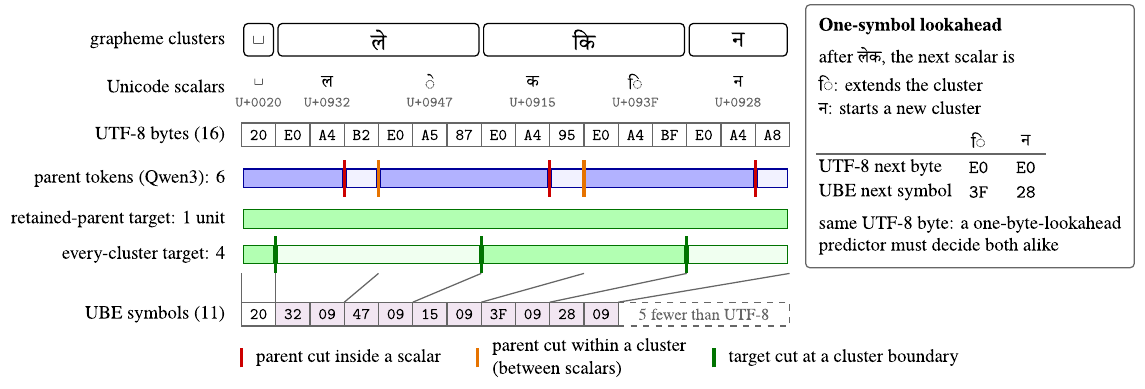}
\caption{Encoding and segmentation of Hindi \emph{lekin} (`but'), tokenized in isolation with a leading space. Qwen3 splits \ExampleBytes{} UTF-8 bytes into \ExampleTokens{} tokens: \ExampleInsideScalar{} cuts fall inside a scalar (red) and \ExampleInsideCluster{} between scalars within a cluster (orange); none falls at a cluster boundary. Deleting these cuts leaves \ExampleSnappedUnits{} target unit; cutting at every cluster boundary gives \ExampleClusterUnits{} units, including the space. UBE uses \ExampleUBESymbols{} symbols (white: UTF-8; violet: UTF-16LE), drawn at the UTF-8 byte width. Inset: after identical left context, the same next UTF-8 byte hides whether the next scalar extends the cluster, whereas UBE's first symbol distinguishes the two (P5). The word was chosen post hoc from the first \ExampleSentences{} Hindi FLORES-200 devtest sentences and is illustrative, not representative.}
\label{fig:example}
\end{figure}

\paragraph{The worked example of Figure~\ref{fig:example}.}
The selection rule, fixed after browsing candidates, chooses the most frequent all-Devanagari word in the first \ExampleSentences{} Hindi devtest sentences that has three to four clusters, excluding a leading space, and whose parent tokenization cuts both inside a scalar and between scalars of one cluster.
Tokenization is performed on the isolated word with that leading space.
Candidate counts use whitespace-split NFC sentences after stripping leading and trailing \texttt{.,;:!?()}, quotation marks, and the Devanagari danda; frequency ties are broken lexicographically.
Tokens come from the \modelid{Qwen/Qwen3-8B} tokenizer at the worked-example revision in Appendix~\ref{app:repro}; clusters use the same \texttt{regex} segmentation as the audit; UBE symbols follow the routing rule in Section~\ref{sec:background}.
Its single retained-parent unit follows from the cuts in this example; cluster alignment does not generally produce one unit per word.

\paragraph{Decision and offset convention.}
For input $[\mathrm{BOS},x_0,\ldots,x_{L-1}]$, hidden state $h_j$ at $j\geq1$ ends at text byte $x_{j-1}$.
For $1\leq j<L$, the predictor compares projected $h_j$ and $h_{j+1}$; an end flag at $j$ therefore denotes a cut at text byte offset $j$, with $x_j$ as lookahead.
The released zero-temperature rule cuts when the boundary log-probability is strictly greater than $\log(1/2)$, so identical visible prefixes receive identical decisions.
It forces the BOS flag at $j=0$ and suppresses the last byte's flag at $j=L$, where lookahead is unavailable.
The audit uses only $1\leq j<L$ as interior cuts and counts the final text unit separately.
The training implementation defaults to end boundaries, although the released configuration does not record which mode was used for training.
The shifted-offset check validates the inference mapping without resolving that missing metadata.
For the UBE argument, transferred vocabulary features must also depend only on the available symbol prefix; using later symbols would change the predictor's information budget.
\FloatBarrier

\section{Boundary-Lookahead Probe}
\label{app:aone}
\paragraph{Design.}
Each of the \AOneCells{} runs trains a predictor with a 513-entry, 128-dimensional symbol embedding (including padding), a one-layer causal LSTM with \AOneHidden{} units, layer normalization, and H-Net's cosine boundary head \citep{hnet} as adapted in Bolmo \citep{bolmo}.
The head places a cut after position $i$ when $(1-\cos(f(h_i),g(h_{i+k})))/2>1/2$, where $f$ and $g$ are learned linear projections.
Training runs for \AOneSteps{} steps with \AOneRows{} rows per step.
Each row begins at a uniformly sampled cluster start in a uniformly sampled language and contains up to \AOneSpan{} supervised scalars, followed by up to three scalars of unsupervised lookahead context.
Row sampling depends only on the seed, ensuring identical text exposure across encodings and lookahead settings within a seed.
The loss adds the mean binary cross-entropy (BCE) over scalar-end decisions to the mean BCE over within-scalar decisions.
We use AdamW with learning rate $2\times10^{-3}$, weight decay 0.01, 100 warmup steps followed by cosine decay to one tenth, and gradient norm clipping at 1; BCE probabilities are clipped to $[10^{-3},1-10^{-3}]$.
The encoder and projections use bfloat16, with the cosine and loss computed in float32.
For each language, training uses whole documents in file order from the first FineWeb-2 shard until the retained NFC text reaches at least \AOneTrainMB{} million UTF-8 bytes.
For each language, documents containing one of that language's FLORES-200 devtest sentences are excluded.
Targets are extended grapheme cluster boundaries from the same \texttt{regex} library version as the boundary audit.
This design matches training text while allowing symbol counts and FLOPs to differ; UTF-8 processes more positions.
The two loss terms have equal aggregate weight, but their gradients can differ across encodings.
UBE-BE preserves UBE's symbol count and loss weights while reordering the entire encoded history, so its contrast with UBE includes effects on history representation and optimization as well as the first lookahead symbol.

\paragraph{Evaluation.}
All conditions are evaluated at the same \AOneSites{} FLORES-200 devtest scalar ends across \AOneLanguages{} languages: \AOneBoundaries{} cluster boundaries and \AOneInside{} scalar ends inside clusters.
Sentence-final boundaries are excluded.
Balanced error averages the missed-boundary and scalar-end false-cut rates after pooling counts across the panel, then averages the two seeds; within-scalar sites are scored separately.
Pooling counts does not give languages equal weight.

Each sentence also supplies a same-context continuation pair.
At a site where a base letter precedes a scalar on the same 256-codepoint page, the true continuation is paired with an NFC-stable replacement from that page requiring the opposite decision.
A pair is correct only when both decisions are correct.
Encoding alone determines the minimum lookahead needed to distinguish them: UTF-8 requires two bytes for \AOnePairsTwoBytes{} pairs and three for the remaining \AOnePairsThreeBytes; UBE requires one symbol for every pair, and UBE-BE requires two.
UTF-8 and UBE-BE therefore have zero fully correct pairs at $k=1$ and exactly 50\% decision error on these pairs by construction.
The trained UBE $k=1$ predictors answer every pair correctly, whereas UTF-8 at $k=2$ answers \AOneUTFTwoPairsLow--\AOneUTFTwoPairsHigh\% of all pairs correctly (\AOneUTFTwoCorrectLow{} and \AOneUTFTwoCorrectHigh{} pairs in its two seeds), although two bytes distinguish \AOnePairsTwoBytes{} pairs.

Contrasts are differences of two-seed means, with 95\% percentile intervals from 2,000 paired source-sentence bootstrap draws that preserve parallel translations.
These pointwise intervals condition on the two trained seeds and include neither multiplicity correction nor training-seed uncertainty.
Table~\ref{tab:aonefull} reports each seed separately to distinguish conditional sentence uncertainty from variation across runs.
The directions of the first four contrasts in Table~\ref{tab:aonecontrasts} were stated before the run.

\begin{table}[htbp]
\caption{Probe errors (\%) and same-context pairs answered correctly, out of \AOnePairs{} per seed. ``Balanced 0 / 1'' gives scalar-end balanced error separately by seed; the other error columns are two-seed means. Missed boundaries and false cuts use only scalar ends, while ``Within scalar'' uses within-scalar sites. Pair counts are also seed 0 / 1.}
\label{tab:aonefull}
\centering\small
\begin{tabular}{lrrrrrr}
\toprule
 &  & \multicolumn{4}{c}{Error rate (\%)} &  \\
\cmidrule(lr){3-6}
Encoding & $k$ & Balanced 0 / 1 & Missed & False cut & Within scalar & Pairs both right \\
\midrule
UTF-8 & 1 & 8.765 / 8.747 & 9.504 & 8.008 & \textbf{0.000} & 0 / 0 \\
UTF-8 & 2 & 6.314 / 6.434 & 5.114 & 7.634 & 0.000 & 4,676 / 4,949 \\
UTF-8 & 3 & 0.217 / 0.179 & 0.049 & 0.347 & 0.000 & 17,204 / 17,204 \\
UBE & 1 & 0.021 / 0.101 & \textbf{0.100} & \textbf{0.021} & 0.194 & 17,204 / 17,204 \\
UBE & 2 & 0.128 / 0.187 & 0.024 & 0.290 & 0.054 & 17,204 / 17,204 \\
UBE & 3 & 0.179 / 0.191 & 0.014 & 0.356 & 0.018 & 17,204 / 17,204 \\
UBE-BE & 1 & 8.421 / 8.810 & 9.199 & 8.032 & \textbf{0.000} & 0 / 0 \\
UBE-BE & 2 & 0.179 / 0.177 & 0.003 & 0.352 & 0.001 & 17,204 / 17,204 \\
UBE-BE & 3 & 0.196 / 0.198 & 0.050 & 0.344 & 0.006 & 17,204 / 17,204 \\
\bottomrule
\end{tabular}

\end{table}

\paragraph{Per-language results.}
Table~\ref{tab:aonelanguage} lists one-symbol balanced error by panel language.
UBE is lowest in every language; Khmer has its largest error and the largest difference between its two seeds.

\begin{table}[htbp]
\caption{One-symbol balanced error (\%) at scalar ends by panel language, two-seed means; the last column lists UBE's two seeds.}
\label{tab:aonelanguage}
\centering\small
\begin{tabular}{llrrrc}
\toprule
 &  & \multicolumn{3}{c}{Balanced error at $k=1$ (\%)} &  \\
\cmidrule(lr){3-5}
Language & Script & UTF-8 & UBE & UBE-BE & UBE seeds 0 / 1 \\
\midrule
Assamese & Bengali & 8.294 & \textbf{0.002} & 7.834 & 0.002 / 0.002 \\
Bengali & Bengali & 8.060 & \textbf{0.005} & 7.350 & 0.005 / 0.005 \\
Burmese & Myanmar & 10.556 & \textbf{0.004} & 10.485 & 0.005 / 0.003 \\
Gujarati & Gujarati & 9.231 & \textbf{0.000} & 8.906 & 0.000 / 0.000 \\
Hindi & Devanagari & 8.504 & \textbf{0.003} & 8.055 & 0.003 / 0.004 \\
Kannada & Kannada & 10.665 & \textbf{0.000} & 10.607 & 0.000 / 0.000 \\
Khmer & Khmer & 5.313 & \textbf{0.803} & 7.991 & 0.159 / 1.447 \\
Malayalam & Malayalam & 8.412 & \textbf{0.086} & 8.108 & 0.067 / 0.105 \\
Marathi & Devanagari & 9.350 & \textbf{0.017} & 8.661 & 0.017 / 0.018 \\
Nepali & Devanagari & 9.170 & \textbf{0.039} & 8.414 & 0.040 / 0.038 \\
Odia & Oriya & 7.906 & \textbf{0.009} & 7.594 & 0.006 / 0.012 \\
Punjabi & Gurmukhi & 10.062 & \textbf{0.001} & 9.488 & 0.000 / 0.002 \\
Shan & Myanmar & 8.060 & \textbf{0.002} & 7.741 & 0.001 / 0.003 \\
Sinhala & Sinhala & 12.359 & \textbf{0.020} & 13.032 & 0.019 / 0.021 \\
Tamil & Tamil & 7.119 & \textbf{0.000} & 6.569 & 0.001 / 0.000 \\
Telugu & Telugu & 8.608 & \textbf{0.018} & 8.448 & 0.016 / 0.020 \\
Thai & Thai & 12.834 & \textbf{0.001} & 12.212 & 0.001 / 0.001 \\
\bottomrule
\end{tabular}

\end{table}

\begin{table}[htbp]
\caption{Probe contrasts in balanced error (percentage points), differences of two-seed means.}
\label{tab:aonecontrasts}
\centering\small
\begin{tabular}{llrc}
\toprule
Contrast & Stated direction & Estimate (pp) & 95\% interval (pp) \\
\midrule
UBE $k{=}1$ $-$ UTF-8 $k{=}1$ & negative & \textminus{}8.695 & [\textminus{}8.757, \textminus{}8.633] \\
UTF-8 $k{=}3$ $-$ UTF-8 $k{=}1$ & negative & \textminus{}8.558 & [\textminus{}8.620, \textminus{}8.496] \\
UBE $k{=}1$ $-$ UBE-BE $k{=}1$ & negative & \textminus{}8.555 & [\textminus{}8.617, \textminus{}8.493] \\
UBE-BE $k{=}2$ $-$ UBE-BE $k{=}1$ & negative & \textminus{}8.438 & [\textminus{}8.499, \textminus{}8.378] \\
UBE $k{=}1$ $-$ UTF-8 $k{=}3$ & none stated & \textminus{}0.137 & [\textminus{}0.148, \textminus{}0.127] \\
\bottomrule
\end{tabular}

\end{table}

\paragraph{Within-scalar cuts.}
UTF-8 predictors make at most \AOneUTFWithinMax{} false cuts in \AOneUTFWithinPositions{} within-scalar positions.
Distinct leading- and continuation-byte patterns provide explicit cues to scalar extent.
UBE assigns both bytes of a UTF-16 code unit to the same symbol range, requiring context to distinguish the two byte positions.
At one-symbol lookahead, UBE places false cuts at \AOneUBEWithin\% of within-scalar sites across its two encoding routes.
The low error of UBE-BE, which shares this symbol range, shows that these errors are not an established unavoidable cost of the encoding.

\paragraph{Joiner sites (post hoc).}
After inspecting the aggregate results, we recomputed every predictor's decisions on the original evaluation device and at the original precision, reproducing all counts exactly.
We then partitioned scalar ends by whether either adjacent scalar is ZERO WIDTH NON-JOINER or ZERO WIDTH JOINER (Table~\ref{tab:aonejoiner}).
The \AOneJoinerSites{} adjacent sites (\AOneJoinerShare\% of scalar ends) occur mainly in Sinhala, Kannada, Malayalam, Odia, and Telugu, including Sinhala yansaya and rakaransaya, whereas the FineWeb-2 training sample contains almost no joiners.
All evaluation scalars lie in the BMP, so UTF-8 at $k=3$ and UBE/UBE-BE at $k\geq2$ see the whole next scalar here.
Away from joiner sites, the maximum balanced error among settings that see the whole next scalar is about \AOneFullOtherMax\%, and UBE reaches \AOneUBEOneOther\% with a single symbol, against \AOneUTFOneOther\% for UTF-8 at one byte.
The fifth contrast, UBE at one symbol minus UTF-8 at three bytes, had no direction specified before the run; its full-panel estimate of $-\AOneGapFifth$ percentage points comes from the joiner-adjacent sites.
The partition is descriptive; the full-panel contrast remains the primary comparison.

\begin{table}[htbp]
\caption{Post hoc: balanced error (\%, two-seed means) at all scalar ends, at the scalar ends next to a zero-width joiner or non-joiner, and elsewhere. Each column uses its own class denominators.}
\label{tab:aonejoiner}
\centering\small
\begin{tabular}{lrrrr}
\toprule
 &  & \multicolumn{3}{c}{Balanced error (\%)} \\
\cmidrule(lr){3-5}
Encoding & $k$ & All scalar ends & Next to a joiner & Other scalar ends \\
\midrule
UTF-8 & 1 & 8.7557 & 50.00 & 8.5994 \\
UTF-8 & 2 & 6.3742 & 50.15 & 6.2133 \\
UTF-8 & 3 & 0.1980 & 61.22 & 0.0019 \\
UBE & 1 & \textbf{0.0609} & \textbf{3.66} & \textbf{0.0492} \\
UBE & 2 & 0.1573 & 40.62 & 0.0123 \\
UBE & 3 & 0.1851 & 53.18 & 0.0013 \\
UBE-BE & 1 & 8.6155 & 86.87 & 8.3972 \\
UBE-BE & 2 & 0.1776 & 49.78 & 0.0008 \\
UBE-BE & 3 & 0.1972 & 59.21 & 0.0054 \\
\bottomrule
\end{tabular}

\end{table}
\FloatBarrier

\section{Korean and Chinese Boundary Extension}
\label{app:cjk}
\paragraph{Scope and targets.}
Designed after the original probe, this exploratory extension examines boundary prediction in Korean and Chinese.
Both targets and all \CJKCells{} runs are reported, with no checkpoint or condition selected by evaluation performance.
The every-cluster target is $G(x)$; the retained-parent target keeps Qwen3 cuts at those cluster boundaries and deletes all other cuts.
The retained target represents inherited subword supervision, not Korean morpheme or Chinese word annotations.
Targets are computed from full-document tokenizations and grapheme segmentation before sampling rows, preserving the original context rather than retokenizing cropped spans.

\paragraph{Data and training.}
For each language, we retain whole documents from the first FineWeb-2 shard until the NFC text reaches at least \CJKTrainMB{} million UTF-8 bytes.
FineWeb-2's Mandarin Han-script source, \texttt{cmn\_Hani}, is not restricted to simplified characters; evaluation uses FLORES-200 Chinese (Simplified), \texttt{zho\_Hans}.
Korean uses \texttt{kor\_Hang} in both corpora.
Any training document containing an evaluation sentence from either language is dropped.
The corpus revision, NFC normalization, grapheme library, and Qwen3 tokenizer match the pinned artifacts reported in Appendix~\ref{app:repro}.
Each predictor has a 513-entry, \CJKEmbed-dimensional embedding, a one-layer \CJKHidden-unit LSTM, and the original probe's cosine head, totaling \CJKParameters{} parameters in every condition.
Training runs for \CJKSteps{} steps with \CJKBatch{} rows per step.
Rows begin at uniformly sampled cluster starts in a uniformly selected language, with up to \CJKSpan{} supervised scalars and three additional scalars for lookahead.
We use float32 throughout and retain the original probe's AdamW optimizer, learning rate, schedule, clipping, and two-term loss.
Text spans are identical across targets, encodings, and lookahead settings within each seed.
The design therefore matches text exposure and parameter counts while allowing symbol counts and FLOPs to differ.
As in the original probe, UBE-BE changes the encoded history as well as the lookahead symbols.
All evaluation scalars lie in the BMP, so UTF-8 at $k=3$ and UBE or UBE-BE at $k=2$ see the whole next scalar on these texts.

\paragraph{Evaluation and uncertainty.}
We evaluate all \CJKSentences{} devtest sentences per language in separate forward passes.
The primary metric sums missed target cuts, false cuts at scalar ends, and false cuts inside scalars, then divides by the original scalar count and multiplies by 1,000.
Sentence-final scalar ends are excluded from the error count; positions inside the final scalar are included.
Using the original scalar count keeps the denominator independent of the number of available cut sites under each encoding.
Supporting missed-boundary and scalar-end false-cut rates (Table~\ref{tab:cjkmetrics}) use their respective positive and negative site counts.
F1 includes within-scalar false cuts, whereas the cluster-split rate counts each affected cluster once.
Reported values average per-seed ratios; the ``Mean'' column in Table~\ref{tab:cjk} then weights the two languages equally.
Contrasts use 2,000 paired source-sentence bootstrap draws, shared across parallel translations, seeds, and conditions.
The 95\% percentile intervals condition on the three trained seeds and are descriptive, without multiplicity correction; Table~\ref{tab:cjkseeds} shows each seed separately.

\begin{table}[htbp]
\caption{Korean/Chinese structural accounting on NFC devtest. ``Retained'' counts target units after deleting incompatible Qwen3 cuts. ``Split'' is the percentage of clusters split by Qwen3, before deletion. Stream savings compare UBE symbols with UTF-8 bytes; UBE-BE has the same length as UBE.}
\label{tab:cjkstructure}
\centering\small
\begin{tabular}{lrrrrrrr}
\toprule
Language & Scalars & UTF-8 & UBE & Saved \% & Parent & Retained & Split \% \\
\midrule
Korean & 65,965 & 157,799 & 111,887 & 29.10 & 45,020 & 41,075 & 4.74 \\
Chinese (Simplified) & 43,248 & 120,803 & 82,081 & 32.05 & 27,701 & 27,391 & 0.61 \\
\bottomrule
\end{tabular}

\end{table}

\paragraph{Deterministic baselines.}
Neither NFC evaluation set contains a multi-scalar cluster, so every evaluated scalar end is a grapheme boundary.
The every-cluster target consequently has no negative scalar-end class, making balanced error undefined.
A deterministic scalar decoder recovers this target without error (Table~\ref{tab:cjkbaselines}), providing a structural baseline rather than a learned predictor with a matched information budget.
For the retained-parent target, the same rule oversegments; predicting no cuts provides the complementary constant-label baseline.

\begin{table}[htbp]
\caption{Deterministic baselines in errors per 1,000 Unicode scalars. Both respect scalar interiors.}
\label{tab:cjkbaselines}
\centering\small
\begin{tabular}{llrr}
\toprule
Target & Baseline & Korean & Chinese \\
\midrule
Every cluster & Scalar decoder & 0.00 & 0.00 \\
Every cluster & No cuts & 984.66 & 976.60 \\
Retained parent cuts & Scalar decoder & 377.32 & 366.65 \\
Retained parent cuts & No cuts & 607.34 & 609.95 \\
\bottomrule
\end{tabular}

\end{table}

\begin{table}[htbp]
\caption{Errors per 1,000 Unicode scalars for every trained seed, including within-scalar cuts.}
\label{tab:cjkseeds}
\centering\small
\begin{tabular}{lrcc}
\toprule
Encoding & $k$ & Korean: seeds 0 / 1 / 2 & Chinese: seeds 0 / 1 / 2 \\
\midrule
\multicolumn{4}{@{}l}{\emph{Every cluster}} \\
UTF-8 & 1 & 0.000 / 0.000 / 0.000 & 0.000 / 0.000 / 0.000 \\
UBE & 1 & 0.000 / 0.030 / 0.076 & 0.023 / 0.000 / 0.116 \\
UBE-BE & 1 & 0.000 / 0.000 / 0.000 & 0.000 / 0.000 / 0.000 \\
UTF-8 & 3 & 0.000 / 0.000 / 0.000 & 0.023 / 0.000 / 0.000 \\
UBE & 2 & 0.258 / 0.015 / 0.015 & 0.069 / 0.000 / 0.000 \\
UBE-BE & 2 & 0.000 / 0.000 / 0.000 & 0.000 / 0.000 / 0.000 \\
\multicolumn{4}{@{}l}{\emph{Retained parent cuts}} \\
UTF-8 & 1 & 53.907 / 56.196 / 53.210 & 160.123 / 157.950 / 161.765 \\
UBE & 1 & 14.174 / 15.417 / 14.644 & 106.132 / 101.693 / 102.664 \\
UBE-BE & 1 & 25.028 / 26.074 / 24.968 & 118.757 / 122.826 / 122.133 \\
UTF-8 & 3 & 10.642 / 12.067 / 10.733 & 101.554 / 102.571 / 105.600 \\
UBE & 2 & 10.142 / 11.203 / 10.324 & 89.900 / 87.334 / 90.594 \\
UBE-BE & 2 & 11.097 / 11.188 / 9.808 & 90.617 / 85.276 / 90.917 \\
\bottomrule
\end{tabular}

\end{table}

\begin{table}[htbp]
\caption{UBE-minus-control contrasts in errors per 1,000 Unicode scalars; negative values favor UBE. Intervals resample parallel source sentences and condition on the three seeds.}
\label{tab:cjkcontrasts}
\centering\small
\begin{tabular}{lrrrr}
\toprule
 & \multicolumn{2}{c}{Korean} & \multicolumn{2}{c}{Chinese} \\
\cmidrule(lr){2-3}\cmidrule(lr){4-5}
Contrast & Difference & 95\% interval & Difference & 95\% interval \\
\midrule
\multicolumn{5}{@{}l}{\emph{Every cluster}} \\
UBE $k=1$ $-$ UTF-8 $k=1$ & +0.035 & [+0.000, +0.095] & +0.046 & [+0.015, +0.086] \\
UBE $k=1$ $-$ UBE-BE $k=1$ & +0.035 & [+0.000, +0.095] & +0.046 & [+0.015, +0.086] \\
UBE $k=2$ $-$ UTF-8 $k=3$ & +0.096 & [+0.005, +0.229] & +0.015 & [+0.000, +0.039] \\
UBE $k=2$ $-$ UBE-BE $k=2$ & +0.096 & [+0.005, +0.229] & +0.023 & [+0.000, +0.062] \\
\multicolumn{5}{@{}l}{\emph{Retained parent cuts}} \\
UBE $k=1$ $-$ UTF-8 $k=1$ & \textminus{}39.693 & [\textminus{}41.276, \textminus{}38.085] & \textminus{}56.450 & [\textminus{}59.506, \textminus{}53.664] \\
UBE $k=1$ $-$ UBE-BE $k=1$ & \textminus{}10.612 & [\textminus{}11.612, \textminus{}9.589] & \textminus{}17.743 & [\textminus{}20.276, \textminus{}15.410] \\
UBE $k=2$ $-$ UTF-8 $k=3$ & \textminus{}0.591 & [\textminus{}1.108, \textminus{}0.092] & \textminus{}13.966 & [\textminus{}15.887, \textminus{}12.186] \\
UBE $k=2$ $-$ UBE-BE $k=2$ & \textminus{}0.141 & [\textminus{}0.551, +0.274] & +0.339 & [\textminus{}1.099, +1.824] \\
\bottomrule
\end{tabular}

\end{table}

\begin{table}[htbp]
\caption{Supporting metrics, averaged over the three seeds. Within-scalar cuts use the common text denominator; missed boundaries and scalar-end false cuts use their own class denominators. A dash means the negative scalar-end class is absent. F1 includes within-scalar false cuts; ``Split'' counts clusters with at least one internal cut.}
\label{tab:cjkmetrics}
\centering\small
\begin{tabular}{lrlrrrrr}
\toprule
Encoding & $k$ & Language & Within / 1,000 & Missed \% & False cut \% & F1 \% & Split \% \\
\midrule
\multicolumn{8}{@{}l}{\emph{Every cluster}} \\
UTF-8 & 1 & Korean & 0.000 & 0.000 & --- & 100.000 & 0.000 \\
UTF-8 & 1 & Chinese & 0.000 & 0.000 & --- & 100.000 & 0.000 \\
UBE & 1 & Korean & 0.025 & 0.001 & --- & 99.998 & 0.003 \\
UBE & 1 & Chinese & 0.031 & 0.002 & --- & 99.998 & 0.003 \\
UBE-BE & 1 & Korean & 0.000 & 0.000 & --- & 100.000 & 0.000 \\
UBE-BE & 1 & Chinese & 0.000 & 0.000 & --- & 100.000 & 0.000 \\
UTF-8 & 3 & Korean & 0.000 & 0.000 & --- & 100.000 & 0.000 \\
UTF-8 & 3 & Chinese & 0.008 & 0.000 & --- & 100.000 & 0.001 \\
UBE & 2 & Korean & 0.045 & 0.005 & --- & 99.995 & 0.005 \\
UBE & 2 & Chinese & 0.008 & 0.002 & --- & 99.999 & 0.001 \\
UBE-BE & 2 & Korean & 0.000 & 0.000 & --- & 100.000 & 0.000 \\
UBE-BE & 2 & Chinese & 0.000 & 0.000 & --- & 100.000 & 0.000 \\
\multicolumn{8}{@{}l}{\emph{Retained parent cuts}} \\
UTF-8 & 1 & Korean & \textbf{0.000} & 5.683 & 5.281 & 95.463 & \textbf{0.000} \\
UTF-8 & 1 & Chinese & 0.000 & 10.036 & 26.928 & 87.279 & 0.000 \\
UBE & 1 & Korean & 0.005 & \textbf{1.800} & \textbf{1.008} & \textbf{98.779} & 0.001 \\
UBE & 1 & Chinese & 0.000 & \textbf{6.357} & \textbf{17.652} & \textbf{91.693} & 0.000 \\
UBE-BE & 1 & Korean & 0.005 & 2.999 & 1.892 & 97.893 & 0.001 \\
UBE-BE & 1 & Chinese & 0.000 & 7.696 & 20.264 & 90.279 & 0.000 \\
UTF-8 & 3 & Korean & 0.005 & 1.299 & 0.862 & 99.079 & 0.001 \\
UTF-8 & 3 & Chinese & 0.015 & 5.743 & 18.600 & 91.761 & 0.002 \\
UBE & 2 & Korean & 0.030 & 1.269 & 0.747 & 99.127 & 0.003 \\
UBE & 2 & Chinese & 0.008 & 5.513 & 15.175 & 92.812 & 0.001 \\
UBE-BE & 2 & Korean & 0.005 & 1.268 & 0.793 & 99.116 & 0.001 \\
UBE-BE & 2 & Chinese & 0.000 & 5.212 & 15.585 & 92.859 & 0.000 \\
\bottomrule
\end{tabular}

\end{table}

\FloatBarrier
\paragraph{Post hoc whitespace partition.}
\phantomsection\label{app:cjkspacing}
We replay all \CJKSpacingRuns{} retained-parent predictors on CPU, first verifying exact agreement with the original aggregate counts for every sentence.
Interior scalar gaps are then partitioned using the recorded Python runtime's Unicode whitespace predicate.
The four disjoint strata have whitespace only to the right (before whitespace), only to the left (after whitespace), on both sides (between whitespace), or on neither side; sentence endpoints are excluded.
Table~\ref{tab:cjkspacingstructure} gives the gap counts and target-cut frequencies.
Whitespace constitutes \CJKKoreanWhitespaceShare\% of Korean scalars and \CJKChineseWhitespaceShare\% of Chinese scalars, so the Chinese text contains whitespace despite its lack of regular word spacing.

Table~\ref{tab:cjkspacing} reports missed cuts plus false cuts as a percentage of each stratum's gaps, averaged across seeds.
Within-scalar errors remain part of the primary metric but lie outside this scalar-gap partition.
The analysis was chosen after the aggregate results and describes errors on the original inputs; it neither removes whitespace nor evaluates linguistic word segmentation.
Every before-whitespace gap is a positive target in these texts, while most after-whitespace gaps are negative, making the distinction between the two sides consequential.
At gaps with no adjacent whitespace, UBE's $k=1$ error is lower than both controls in every seed of each language.
Chinese gaps immediately after whitespace show the opposite pattern, favoring both UTF-8 and UBE-BE over UBE.

\begin{table}[htbp]
\caption{Whitespace partition of the retained-parent target. Counts refer to interior scalar gaps; target-cut percentages use the row's gap count. A dash marks an absent stratum.}
\label{tab:cjkspacingstructure}
\centering\small
\begin{tabular}{llrr}
\toprule
Language & Position & Scalar gaps & Target cuts \% \\
\midrule
Korean & Before whitespace & 14,418 & 100.00 \\
Korean & After whitespace & 14,418 & 2.53 \\
Korean & Between whitespace & 4 & 75.00 \\
Korean & Neither side & 36,113 & 69.99 \\
Chinese & Before whitespace & 1,057 & 100.00 \\
Chinese & After whitespace & 1,057 & 36.23 \\
Chinese & Between whitespace & 0 & --- \\
Chinese & Neither side & 40,122 & 62.16 \\
\bottomrule
\end{tabular}

\end{table}

\begin{table}[htbp]
\caption{Post hoc scalar-gap error percentages by whitespace context. ``Before,'' ``After,'' and ``Between'' refer to whitespace; ``Neither side'' has no whitespace on either side. All encoding/lookahead conditions are included, with \CJKSeeds-seed means. These percentages have different denominators from the primary errors per 1,000 scalars.}
\label{tab:cjkspacing}
\centering\small
\begin{tabular}{lrrrrr}
\toprule
Encoding & $k$ & Before & After & Between & Neither side \\
\midrule
\multicolumn{6}{@{}l}{\emph{Korean}} \\
UTF-8 & 1 & 0.000 & \textbf{0.000} & \textbf{25.000} & 9.941 \\
UBE & 1 & 0.000 & 0.005 & \textbf{25.000} & \textbf{2.688} \\
UBE-BE & 1 & 0.000 & \textbf{0.000} & 33.333 & 4.627 \\
UTF-8 & 3 & 0.009 & 0.000 & 75.000 & 2.023 \\
UBE & 2 & 0.000 & 0.000 & 25.000 & 1.920 \\
UBE-BE & 2 & 0.000 & 0.000 & 41.667 & 1.949 \\
\multicolumn{6}{@{}l}{\emph{Chinese}} \\
UTF-8 & 1 & 0.000 & 4.793 & --- & 17.114 \\
UBE & 1 & 0.000 & 8.704 & --- & \textbf{10.927} \\
UBE-BE & 1 & 0.000 & \textbf{2.838} & --- & 12.994 \\
UTF-8 & 3 & 0.000 & 0.158 & --- & 11.123 \\
UBE & 2 & 0.000 & 0.315 & --- & 9.614 \\
UBE-BE & 2 & 0.000 & 0.473 & --- & 9.574 \\
\bottomrule
\end{tabular}

\end{table}
\FloatBarrier

\section{Eight More Scripts and Learning Curves}
\label{app:scriptext}
\paragraph{Design.}
This exploratory extension repeats the retained-parent comparison of Appendix~\ref{app:cjk} in eight more languages: Japanese (\texttt{jpn\_Jpan}), Thai (\texttt{tha\_Thai}), Lao (\texttt{lao\_Laoo}), Khmer (\texttt{khm\_Khmr}), Burmese (\texttt{mya\_Mymr}), Hindi (\texttt{hin\_Deva}), Bengali (\texttt{ben\_Beng}), and Tamil (\texttt{tam\_Taml}).
Each predictor is trained on one language only, with the model, optimizer, \CJKSteps{} steps of \CJKBatch{} rows, six encoding and lookahead conditions, and \CJKSeeds{} seeds of Appendix~\ref{app:cjk}; text spans are identical across conditions within each language and seed.
For each language, we retain whole documents from the first FineWeb-2 shard until the NFC text reaches at least \CJKTrainMB{} million UTF-8 bytes; no retained document contains a FLORES-200 devtest sentence of any of the eight languages.
Targets keep the Qwen3 cuts that fall on extended grapheme cluster boundaries and delete all others, so outside Japanese they also remove cuts inside multi-scalar clusters.
We evaluate all devtest sentences of each language after every \XThreeCheckpointEvery{} updates and report all \XOneCells{} runs.
Intervals use \XOneBootstrapDraws{} paired source-sentence bootstrap draws shared across seeds and conditions and are conditional on the \CJKSeeds{} seeds.
All evaluation scalars lie in the BMP, so UTF-8 at $k=3$ and UBE or UBE-BE at $k=2$ see the whole next scalar.

\paragraph{Pages and first symbols.}
The main blocks of Thai (U+0E00--U+0E7F), Lao (U+0E80--U+0EFF), Khmer (U+1780--U+17FF), Myanmar (U+1000--U+109F), Devanagari (U+0900--U+097F), Bengali (U+0980--U+09FF), and Tamil (U+0B80--U+0BFF) each lie on a single 256-codepoint page \citep{unicode2025}.
Every letter of such a block therefore has the same first UTF-8 byte (\texttt{E0} or \texttt{E1}) and the same first UBE-BE symbol, its page, whereas its first UBE symbol is its position within the page.
Japanese text mixes hiragana and katakana from their main blocks, which share one page, with kanji spread over many, so UBE-BE's first symbol separates these kana from kanji and distinguishes kanji pages.
The texts also contain Latin letters, digits, and punctuation, so a constant first symbol within one block does not by itself determine the learned error.

\paragraph{Results.}
Table~\ref{tab:scriptextcontrasts} gives the paired contrasts.
UBE at $k=1$ has fewer errors than both UTF-8 and UBE-BE in every seed of every language, and every interval for these two contrasts excludes zero.
In the single-page scripts, UBE-BE differs from UTF-8 by at most \XOneBEMaxGap\% at $k=1$, and UBE's reduction relative to UBE-BE is at least \XOneMinBEShare\% of its reduction relative to UTF-8.
In Japanese, UBE-BE lies between UTF-8 and UBE.
Table~\ref{tab:scriptextcurves} lists the seed-mean errors at every checkpoint.

\paragraph{Korean and Chinese learning curves.}
To obtain learning curves for the Korean/Chinese predictors, we retrained the \XThreeCells{} retained-parent runs of Appendix~\ref{app:cjk} with checkpoint evaluations.
Every retrained model has the same text spans, bitwise identical weights, and the same final counts as the original run, so the Korean and Chinese curves in Table~\ref{tab:scriptextcurves} belong to the predictors reported in Table~\ref{tab:cjk}.

\begin{table*}[htbp]
\caption{Paired contrasts for the eight-script extension: differences in errors per 1,000 scalars between three-seed means, each with its 95\% source-sentence bootstrap interval in brackets below. Negative values favor the first encoding. The whole-next-scalar columns compare UBE at $k=2$ with UTF-8 at $k=3$ and with UBE-BE at $k=2$.}
\label{tab:scriptextcontrasts}
\centering\small
\begin{tabular}{lccccc}
\toprule
 & \multicolumn{3}{c}{One symbol ($k=1$)} & \multicolumn{2}{c}{Whole next scalar} \\
\cmidrule(lr){2-4}\cmidrule(lr){5-6}
Language & UBE $-$ UTF-8 & UBE $-$ UBE-BE & UBE-BE $-$ UTF-8 & UBE $-$ UTF-8 & UBE $-$ UBE-BE \\
\midrule
Japanese & \textminus{}79.94 & \textminus{}71.05 & \textminus{}8.89 & \textminus{}2.53 & \textminus{}0.17 \\
 & [\textminus{}82.55, \textminus{}77.43] & [\textminus{}73.64, \textminus{}68.52] & [\textminus{}10.33, \textminus{}7.42] & [\textminus{}3.37, \textminus{}1.66] & [\textminus{}0.89, +0.60] \\
Thai & \textminus{}56.37 & \textminus{}54.73 & \textminus{}1.64 & \textminus{}0.67 & \textminus{}0.22 \\
 & [\textminus{}58.10, \textminus{}54.64] & [\textminus{}56.42, \textminus{}53.01] & [\textminus{}2.18, \textminus{}1.08] & [\textminus{}0.98, \textminus{}0.34] & [\textminus{}0.51, +0.08] \\
Lao & \textminus{}67.00 & \textminus{}65.83 & \textminus{}1.18 & +0.18 & \textminus{}0.02 \\
 & [\textminus{}68.47, \textminus{}65.54] & [\textminus{}67.28, \textminus{}64.35] & [\textminus{}1.80, \textminus{}0.57] & [\textminus{}0.03, +0.38] & [\textminus{}0.17, +0.13] \\
Khmer & \textminus{}44.29 & \textminus{}45.45 & +1.17 & +0.04 & +0.23 \\
 & [\textminus{}45.46, \textminus{}43.15] & [\textminus{}46.57, \textminus{}44.30] & [+0.69, +1.64] & [\textminus{}0.14, +0.22] & [+0.09, +0.37] \\
Burmese & \textminus{}101.70 & \textminus{}98.81 & \textminus{}2.89 & +0.15 & +0.02 \\
 & [\textminus{}103.15, \textminus{}100.32] & [\textminus{}100.18, \textminus{}97.36] & [\textminus{}3.52, \textminus{}2.26] & [+0.04, +0.27] & [\textminus{}0.05, +0.11] \\
Hindi & \textminus{}66.94 & \textminus{}66.76 & \textminus{}0.19 & \textminus{}0.04 & \textminus{}0.02 \\
 & [\textminus{}68.22, \textminus{}65.68] & [\textminus{}68.07, \textminus{}65.45] & [\textminus{}0.70, +0.31] & [\textminus{}0.13, +0.04] & [\textminus{}0.07, +0.02] \\
Bengali & \textminus{}79.00 & \textminus{}76.65 & \textminus{}2.35 & \textminus{}0.50 & \textminus{}0.76 \\
 & [\textminus{}80.62, \textminus{}77.55] & [\textminus{}78.23, \textminus{}75.17] & [\textminus{}2.88, \textminus{}1.82] & [\textminus{}0.68, \textminus{}0.32] & [\textminus{}0.92, \textminus{}0.60] \\
Tamil & \textminus{}69.28 & \textminus{}68.38 & \textminus{}0.90 & \textminus{}0.25 & \textminus{}0.31 \\
 & [\textminus{}70.56, \textminus{}68.01] & [\textminus{}69.63, \textminus{}67.08] & [\textminus{}1.34, \textminus{}0.49] & [\textminus{}0.46, +0.02] & [\textminus{}0.52, \textminus{}0.04] \\
\bottomrule
\end{tabular}

\end{table*}

\begin{table*}[htbp]
\caption{Learning curves at one-symbol lookahead ($k=1$): retained-parent errors per 1,000 scalars after each number of updates, averaged over three seeds. The final column equals the $k=1$ values in Tables~\ref{tab:cjk} and~\ref{tab:scriptext}.}
\label{tab:scriptextcurves}
\centering\small
\begin{tabular}{llrrrrrrrr}
\toprule
Language & Encoding & 250 & 500 & 750 & 1,000 & 1,250 & 1,500 & 1,750 & 2,000 \\
\midrule
\multicolumn{10}{@{}l}{\emph{Korean and Chinese (Section~\ref{sec:cjk})}} \\
Korean & UTF-8 & 80.1 & 64.1 & 62.0 & 57.9 & 58.4 & 54.8 & 54.2 & 54.4 \\
Korean & UBE-BE & 60.6 & 37.8 & 31.2 & 29.3 & 27.0 & 26.4 & 25.7 & 25.4 \\
Korean & UBE & 62.1 & 30.2 & 21.6 & 18.5 & 16.0 & 15.6 & 15.1 & 14.7 \\
Chinese & UTF-8 & 235.7 & 210.3 & 192.6 & 180.4 & 173.0 & 164.9 & 162.2 & 159.9 \\
Chinese & UBE-BE & 215.8 & 181.7 & 158.6 & 143.6 & 135.9 & 127.7 & 123.4 & 121.2 \\
Chinese & UBE & 217.6 & 173.2 & 145.6 & 131.3 & 119.4 & 110.7 & 107.2 & 103.5 \\
\multicolumn{10}{@{}l}{\emph{Eight more scripts (Section~\ref{sec:scripts})}} \\
Japanese & UTF-8 & 172.3 & 149.2 & 136.7 & 129.7 & 125.6 & 122.3 & 120.6 & 118.7 \\
Japanese & UBE-BE & 163.4 & 140.0 & 127.8 & 119.9 & 115.2 & 112.6 & 110.3 & 109.8 \\
Japanese & UBE & 105.1 & 67.1 & 54.2 & 47.6 & 44.7 & 41.3 & 39.3 & 38.8 \\
Thai & UTF-8 & 111.4 & 96.3 & 87.5 & 83.7 & 80.2 & 78.3 & 76.5 & 76.2 \\
Thai & UBE-BE & 107.9 & 92.5 & 85.0 & 81.0 & 78.2 & 76.5 & 74.9 & 74.5 \\
Thai & UBE & 41.6 & 30.0 & 24.8 & 22.1 & 21.1 & 20.3 & 19.6 & 19.8 \\
Lao & UTF-8 & 111.2 & 95.6 & 85.1 & 80.0 & 77.3 & 75.9 & 73.9 & 73.3 \\
Lao & UBE-BE & 108.7 & 92.1 & 84.4 & 79.0 & 76.3 & 74.5 & 73.5 & 72.2 \\
Lao & UBE & 6.7 & 6.6 & 6.5 & 6.6 & 6.6 & 6.5 & 6.4 & 6.3 \\
Khmer & UTF-8 & 90.8 & 68.7 & 60.7 & 57.7 & 55.9 & 55.0 & 55.1 & 53.6 \\
Khmer & UBE-BE & 96.4 & 75.3 & 62.7 & 62.4 & 57.5 & 58.2 & 55.8 & 54.8 \\
Khmer & UBE & 10.4 & 9.5 & 9.6 & 10.0 & 9.4 & 9.2 & 9.3 & 9.3 \\
Burmese & UTF-8 & 139.2 & 125.0 & 118.9 & 113.0 & 108.9 & 106.4 & 104.6 & 103.1 \\
Burmese & UBE-BE & 133.8 & 120.6 & 113.6 & 109.4 & 105.7 & 102.9 & 101.3 & 100.2 \\
Burmese & UBE & 2.7 & 2.3 & 2.3 & 1.8 & 1.5 & 1.4 & 1.4 & 1.4 \\
Hindi & UTF-8 & 95.4 & 86.5 & 80.1 & 76.2 & 71.7 & 69.5 & 68.9 & 67.2 \\
Hindi & UBE-BE & 93.7 & 84.6 & 78.5 & 74.5 & 71.4 & 69.3 & 68.6 & 67.1 \\
Hindi & UBE & 0.8 & 0.5 & 0.4 & 0.4 & 0.4 & 0.3 & 0.3 & 0.3 \\
Bengali & UTF-8 & 105.4 & 96.1 & 89.9 & 87.0 & 83.3 & 82.1 & 80.8 & 80.3 \\
Bengali & UBE-BE & 103.7 & 93.4 & 87.5 & 84.2 & 81.3 & 80.0 & 79.2 & 78.0 \\
Bengali & UBE & 2.8 & 2.3 & 2.1 & 1.8 & 1.5 & 1.5 & 1.4 & 1.3 \\
Tamil & UTF-8 & 101.5 & 89.8 & 83.8 & 78.2 & 74.6 & 72.5 & 70.7 & 70.2 \\
Tamil & UBE-BE & 98.2 & 87.4 & 80.9 & 76.3 & 72.7 & 71.2 & 69.9 & 69.3 \\
Tamil & UBE & 1.7 & 1.2 & 1.0 & 1.0 & 1.0 & 0.9 & 1.0 & 1.0 \\
\bottomrule
\end{tabular}

\end{table*}
\FloatBarrier

\section{Full Behavioral Contrasts and Sensitivity}
\label{app:probe}
Tables~\ref{tab:probeone} and \ref{tab:probetwo} report both parent/conversion pairs, each evaluated on its own matched item bank and prefix convention.
Mark and base margins are increases in path cost relative to the original text, averaged in bits; $R$ is the mean mark margin divided by the mean base margin.
$\Delta R$ and $\Delta I$ denote conversion-minus-parent differences.
All mean base margins are positive, so larger $R$ indicates larger relative mark penalties.
Smaller $I$ indicates a shift toward mark penalties under item-level normalization; the two summaries use different weights and need not move together.

\begin{table}[htbp]
\caption{Qwen3-8B-Base and Bwen-8B, \SourceSentences{} items per language. Paired, pointwise percentile intervals; no multiplicity correction.}
\label{tab:probeone}
\centering\footnotesize\setlength{\tabcolsep}{2.6pt}
\begin{tabular}{lrrrrrrrcrc}
\toprule
 & \multicolumn{2}{c}{Parent margin} & \multicolumn{2}{c}{Conv.\ margin} & \multicolumn{2}{c}{Ratio $R$} &  &  &  &  \\
\cmidrule(lr){2-3}\cmidrule(lr){4-5}\cmidrule(lr){6-7}
Language & base & mark & base & mark & parent & conv. & $\Delta R$ & 95\% interval & $\Delta I$ & 95\% interval \\
\midrule
Assamese & 25.88 & 22.49 & 19.57 & 16.35 & 0.8689 & 0.8354 & \textminus{}0.0335 & [\textminus{}0.0865, +0.0197] & \textminus{}0.0123 & [\textminus{}0.0527, +0.0261] \\
Bengali & 29.69 & 25.90 & 25.41 & 22.35 & 0.8722 & 0.8796 & +0.0074 & [\textminus{}0.0358, +0.0506] & \textminus{}0.0083 & [\textminus{}0.0350, +0.0175] \\
Gujarati & 28.52 & 22.37 & 24.41 & 19.96 & 0.7845 & 0.8179 & +0.0334 & [+0.0059, +0.0636] & \textminus{}0.0260 & [\textminus{}0.0475, \textminus{}0.0049] \\
Hindi & 30.26 & 24.51 & 27.92 & 22.99 & 0.8098 & 0.8232 & +0.0135 & [\textminus{}0.0147, +0.0415] & \textminus{}0.0126 & [\textminus{}0.0317, +0.0056] \\
Kannada & 30.51 & 24.70 & 24.70 & 20.78 & 0.8095 & 0.8412 & +0.0317 & [\textminus{}0.0067, +0.0715] & \textminus{}0.0184 & [\textminus{}0.0460, +0.0099] \\
Khmer & 28.48 & 26.91 & 20.09 & 18.76 & 0.9446 & 0.9337 & \textminus{}0.0109 & [\textminus{}0.0651, +0.0438] & +0.0088 & [\textminus{}0.0323, +0.0505] \\
Malayalam & 31.93 & 24.94 & 24.63 & 20.33 & 0.7812 & 0.8254 & +0.0441 & [+0.0071, +0.0831] & \textminus{}0.0273 & [\textminus{}0.0552, +0.0009] \\
Marathi & 28.25 & 23.84 & 25.00 & 20.92 & 0.8439 & 0.8366 & \textminus{}0.0072 & [\textminus{}0.0486, +0.0362] & \textminus{}0.0060 & [\textminus{}0.0336, +0.0215] \\
Burmese & 29.24 & 27.34 & 19.63 & 21.73 & 0.9349 & 1.1070 & +0.1720 & [+0.1222, +0.2233] & \textminus{}0.1010 & [\textminus{}0.1319, \textminus{}0.0726] \\
Nepali & 25.92 & 21.58 & 23.01 & 19.46 & 0.8324 & 0.8458 & +0.0133 & [\textminus{}0.0304, +0.0639] & +0.0112 & [\textminus{}0.0171, +0.0391] \\
Odia & 26.68 & 22.59 & 22.51 & 17.98 & 0.8468 & 0.7986 & \textminus{}0.0482 & [\textminus{}0.0856, \textminus{}0.0114] & +0.0213 & [\textminus{}0.0065, +0.0496] \\
Punjabi & 28.74 & 25.06 & 24.52 & 19.80 & 0.8718 & 0.8075 & \textminus{}0.0643 & [\textminus{}0.1051, \textminus{}0.0237] & +0.0320 & [+0.0055, +0.0579] \\
Shan & 10.98 & 15.72 & 4.95 & 10.46 & 1.4312 & 2.1123 & +0.6811 & [+0.3559, +1.1683] & \textminus{}0.1377 & [\textminus{}0.2090, \textminus{}0.0623] \\
Sinhala & 26.57 & 19.77 & 20.65 & 15.19 & 0.7441 & 0.7355 & \textminus{}0.0086 & [\textminus{}0.0523, +0.0362] & \textminus{}0.0026 & [\textminus{}0.0453, +0.0388] \\
Tamil & 30.57 & 26.80 & 25.15 & 21.92 & 0.8767 & 0.8716 & \textminus{}0.0051 & [\textminus{}0.0439, +0.0351] & +0.0069 & [\textminus{}0.0183, +0.0324] \\
Telugu & 28.87 & 23.17 & 24.18 & 19.39 & 0.8026 & 0.8018 & \textminus{}0.0008 & [\textminus{}0.0340, +0.0350] & \textminus{}0.0004 & [\textminus{}0.0281, +0.0275] \\
Thai & 34.62 & 28.63 & 25.68 & 24.68 & 0.8271 & 0.9609 & +0.1338 & [+0.0680, +0.1999] & \textminus{}0.0694 & [\textminus{}0.1105, \textminus{}0.0280] \\
\bottomrule
\end{tabular}

\end{table}

\begin{table}[htbp]
\caption{OLMo-2-0425-1B and Bolmo-1B, \SourceSentences{} items per language. This bank uses the OLMo parent tokenizer convention. Its items and prefixes need not equal those of the first pair.}
\label{tab:probetwo}
\centering\footnotesize\setlength{\tabcolsep}{2.6pt}
\begin{tabular}{lrrrrrrrcrc}
\toprule
 & \multicolumn{2}{c}{Parent margin} & \multicolumn{2}{c}{Conv.\ margin} & \multicolumn{2}{c}{Ratio $R$} &  &  &  &  \\
\cmidrule(lr){2-3}\cmidrule(lr){4-5}\cmidrule(lr){6-7}
Language & base & mark & base & mark & parent & conv. & $\Delta R$ & 95\% interval & $\Delta I$ & 95\% interval \\
\midrule
Assamese & 17.01 & 16.62 & 11.98 & 11.35 & 0.9772 & 0.9471 & \textminus{}0.0301 & [\textminus{}0.1002, +0.0424] & +0.0358 & [\textminus{}0.0160, +0.0888] \\
Bengali & 21.17 & 19.94 & 15.78 & 15.39 & 0.9421 & 0.9753 & +0.0331 & [\textminus{}0.0190, +0.0872] & \textminus{}0.0282 & [\textminus{}0.0613, +0.0045] \\
Gujarati & 20.17 & 16.66 & 13.52 & 11.86 & 0.8257 & 0.8775 & +0.0517 & [\textminus{}0.0022, +0.1114] & \textminus{}0.0578 & [\textminus{}0.1023, \textminus{}0.0131] \\
Hindi & 24.57 & 22.18 & 21.42 & 19.01 & 0.9029 & 0.8873 & \textminus{}0.0156 & [\textminus{}0.0558, +0.0252] & \textminus{}0.0129 & [\textminus{}0.0462, +0.0189] \\
Kannada & 22.32 & 18.32 & 15.94 & 12.69 & 0.8205 & 0.7965 & \textminus{}0.0240 & [\textminus{}0.0655, +0.0180] & +0.0315 & [\textminus{}0.0018, +0.0688] \\
Khmer & 20.54 & 20.77 & 13.61 & 14.52 & 1.0110 & 1.0663 & +0.0553 & [\textminus{}0.0097, +0.1176] & \textminus{}0.0160 & [\textminus{}0.0588, +0.0282] \\
Malayalam & 21.47 & 17.40 & 14.94 & 11.59 & 0.8102 & 0.7760 & \textminus{}0.0342 & [\textminus{}0.0805, +0.0123] & +0.0124 & [\textminus{}0.0273, +0.0551] \\
Marathi & 20.04 & 18.36 & 15.09 & 15.59 & 0.9160 & 1.0328 & +0.1168 & [+0.0509, +0.1850] & \textminus{}0.0574 & [\textminus{}0.0975, \textminus{}0.0183] \\
Burmese & 12.81 & 14.70 & 9.99 & 11.81 & 1.1469 & 1.1820 & +0.0351 & [\textminus{}0.0311, +0.1061] & \textminus{}0.0029 & [\textminus{}0.0411, +0.0371] \\
Nepali & 16.95 & 16.38 & 15.64 & 14.81 & 0.9667 & 0.9466 & \textminus{}0.0201 & [\textminus{}0.0771, +0.0395] & +0.0121 & [\textminus{}0.0296, +0.0545] \\
Odia & 17.78 & 16.73 & 9.22 & 7.87 & 0.9413 & 0.8532 & \textminus{}0.0882 & [\textminus{}0.1756, +0.0016] & +0.0687 & [+0.0068, +0.1282] \\
Punjabi & 19.28 & 17.20 & 12.01 & 10.60 & 0.8920 & 0.8821 & \textminus{}0.0098 & [\textminus{}0.0714, +0.0542] & \textminus{}0.0306 & [\textminus{}0.0837, +0.0237] \\
Shan & 5.32 & 7.57 & 4.99 & 7.65 & 1.4241 & 1.5336 & +0.1095 & [\textminus{}0.0823, +0.3288] & \textminus{}0.0075 & [\textminus{}0.0798, +0.0663] \\
Sinhala & 15.13 & 12.10 & 12.53 & 10.46 & 0.7994 & 0.8347 & +0.0352 & [\textminus{}0.0196, +0.0904] & \textminus{}0.0311 & [\textminus{}0.0792, +0.0163] \\
Tamil & 21.78 & 19.04 & 16.33 & 13.56 & 0.8742 & 0.8304 & \textminus{}0.0438 & [\textminus{}0.0928, +0.0069] & \textminus{}0.0030 & [\textminus{}0.0438, +0.0393] \\
Telugu & 19.59 & 16.70 & 14.25 & 12.59 & 0.8522 & 0.8834 & +0.0312 & [\textminus{}0.0195, +0.0857] & \textminus{}0.0324 & [\textminus{}0.0791, +0.0114] \\
Thai & 23.13 & 28.07 & 18.70 & 21.08 & 1.2135 & 1.1270 & \textminus{}0.0866 & [\textminus{}0.1434, \textminus{}0.0250] & +0.0379 & [+0.0019, +0.0742] \\
\bottomrule
\end{tabular}

\end{table}

\paragraph{Intervals and weighting.}
Panel intervals use \BootstrapDraws{} paired percentile bootstrap draws of the \SourceSentences{} source sentences, retaining all \ProbeLanguages{} translations and both models together.
Per-language intervals instead resample that language's items, again keeping models paired.
The panel ratio pools margins before division, implicitly weighting languages by each model's mean base margins.
It therefore differs from both an unweighted mean of language ratios and an average of language differences under common weights.
The item contrast gives equal weight to each defined item.
Both intervals remain conditional on the selected checkpoints and languages.

\begin{table}[htbp]
\caption{Leave-one-language-out point estimates for pair one.}
\label{tab:omissions}
\centering\small
\begin{tabular}{lrr}
\toprule
Omitted language & $\Delta R$ & $\Delta I$ \\
\midrule
Assamese & +0.0205 & \textminus{}0.0206 \\
Bengali & +0.0183 & \textminus{}0.0208 \\
Gujarati & +0.0169 & \textminus{}0.0197 \\
Hindi & +0.0185 & \textminus{}0.0206 \\
Kannada & +0.0167 & \textminus{}0.0202 \\
Khmer & +0.0200 & \textminus{}0.0219 \\
Malayalam & +0.0156 & \textminus{}0.0197 \\
Marathi & +0.0195 & \textminus{}0.0210 \\
Burmese & +0.0102 & \textminus{}0.0150 \\
Nepali & +0.0181 & \textminus{}0.0221 \\
Odia & +0.0218 & \textminus{}0.0227 \\
Punjabi & +0.0232 & \textminus{}0.0234 \\
Shan & +0.0150 & \textminus{}0.0128 \\
Sinhala & +0.0190 & \textminus{}0.0212 \\
Tamil & +0.0192 & \textminus{}0.0218 \\
Telugu & +0.0191 & \textminus{}0.0213 \\
Thai & +0.0091 & \textminus{}0.0170 \\
\bottomrule
\end{tabular}

\end{table}
\FloatBarrier

\section{Instability Across Checkpoints}
\label{app:instability}
Table~\ref{tab:instabilityfull} uses the first \SubsetPerLanguage{} items per language from each diagnostic bank.
Bwen's items form a subset of its larger behavioral probe; their aggregate identity hash matches the flag diagnostic's bank hash, and their path costs match the stored probe costs exactly.
The two dolma2 conversions share item identities and scored byte windows.
Corruption instability is measured only with standalone forward passes; batched forwards serve a separate diagnostic.

\begin{table}[htbp]
\caption{Per-language instability, agreement of the dolma2 pair's instability indicators (1B=7B, out of \SubsetPerLanguage{} items), flag share of path cost, and bytes per patch. Bwen uses a different item bank, precluding a paired magnitude comparison with the dolma2 models.}
\label{tab:instabilityfull}
\centering\footnotesize\setlength{\tabcolsep}{3pt}
\begin{tabular}{lrrrrrrrrrr}
\toprule
 & \multicolumn{3}{c}{Instability (\%)} & \multicolumn{1}{c}{of 50} & \multicolumn{3}{c}{Flag share (\%)} & \multicolumn{3}{c}{Bytes per patch} \\
\cmidrule(lr){2-4}\cmidrule(lr){5-5}\cmidrule(lr){6-8}\cmidrule(lr){9-11}
Language & Bwen & 1B & 7B & 1B=7B & Bwen & 1B & 7B & Bwen & 1B & 7B \\
\midrule
Assamese & 72 & 64 & 64 & 50 & 8.97 & 13.07 & 13.18 & 2.352 & 2.057 & 2.053 \\
Bengali & 74 & 82 & 82 & 50 & 7.50 & 14.20 & 14.16 & 2.523 & 2.201 & 2.200 \\
Gujarati & 34 & 0 & 0 & 50 & 12.48 & 1.21 & 1.36 & 1.764 & 1.590 & 1.591 \\
Hindi & 68 & 80 & 80 & 50 & 15.64 & 17.82 & 18.49 & 2.771 & 2.629 & 2.630 \\
Kannada & 42 & 0 & 0 & 50 & 13.37 & 1.37 & 2.00 & 1.950 & 1.566 & 1.564 \\
Khmer & 42 & 32 & 32 & 50 & 11.42 & 10.86 & 11.35 & 2.456 & 1.859 & 1.860 \\
Malayalam & 38 & 10 & 8 & 49 & 17.33 & 7.96 & 7.90 & 2.054 & 1.701 & 1.700 \\
Marathi & 60 & 70 & 70 & 50 & 12.66 & 16.46 & 17.09 & 2.784 & 2.605 & 2.605 \\
Burmese & 18 & 0 & 0 & 50 & 15.56 & 4.20 & 4.05 & 1.889 & 1.473 & 1.473 \\
Nepali & 68 & 82 & 82 & 50 & 13.39 & 16.68 & 17.16 & 2.795 & 2.631 & 2.632 \\
Odia & 66 & 16 & 16 & 50 & 22.62 & 7.13 & 8.21 & 1.360 & 1.075 & 1.075 \\
Punjabi & 22 & 0 & 0 & 50 & 9.61 & 1.85 & 2.67 & 1.724 & 1.615 & 1.615 \\
Shan & 64 & 66 & 64 & 49 & 15.40 & 19.30 & 19.91 & 1.807 & 1.284 & 1.282 \\
Sinhala & 46 & 4 & 4 & 50 & 8.87 & 3.19 & 3.35 & 1.814 & 1.469 & 1.469 \\
Tamil & 72 & 70 & 70 & 50 & 13.48 & 13.58 & 13.33 & 2.502 & 2.047 & 2.047 \\
Telugu & 42 & 2 & 0 & 49 & 13.21 & 1.93 & 1.94 & 1.823 & 1.573 & 1.574 \\
Thai & 88 & 78 & 80 & 45 & 9.58 & 8.31 & 8.88 & 4.951 & 3.027 & 3.060 \\
\bottomrule
\end{tabular}

\end{table}

\begin{table}[htbp]
\caption{Unweighted Pearson associations over language aggregates. They do not identify a mechanism or propagate uncertainty in the aggregates.}
\label{tab:associations}
\centering\small
\begin{tabular}{lrrr}
\toprule
Model & Bytes/patch, instability & Bytes/patch, flag share & Flag share, instability \\
\midrule
Bwen-8B & +0.620 & \textminus{}0.369 & \textminus{}0.091 \\
Bolmo-1B & +0.757 & +0.511 & +0.894 \\
Bolmo-7B & +0.764 & +0.504 & +0.883 \\
\bottomrule
\end{tabular}

\end{table}

The two model sizes also differ in pretrained backbone and training history, so agreement supports recurrence within this conversion family without ruling out model-specific effects.
Flag shares are measured only on original sentences; explaining corruption margins would additionally require flag measurements on the edited texts.
\FloatBarrier

\section{Scoring Protocol and Diagnostics}
\label{app:scoring}
Let $t_0$ be the first ordinary parent token and $n_0$ the number of bytes recovered from its byte-alphabet representation.
The scorer verifies that concatenating the bytes of all parent pieces reconstructs the exact NFC sentence.
Recovering piece bytes avoids decoding $t_0$ in isolation, which can replace an incomplete scalar with a replacement character, because byte-level pieces need not be valid UTF-8 \citep{firestone2025illformed}.
Each corruption preserves $t_0$, ensuring that the scored suffix begins at the same byte offset across variants within a pair.

The parent cost is $-\log_2 P(t_1,\ldots,t_{n-1}\mid t_0)$, without a prepended BOS or appended EOS.
The conversion receives its BOS and prefix bytes under the boundary history selected during prefill; the first $n_0$ byte targets are excluded from the score.
The models therefore score the same text suffix under different conditioning representations.
With one-byte lookahead, the final unscored prefix flag can depend on the first scored byte.
This is a byte-level analogue of the prompt boundary problem of tokenized models \citep{vieira2025characters,hayase2026bytesampler}.
In the Bwen bank, \PrefixFlagChanges{} of \PrefixCorruptions{} corruptions change the recorded prefix flags.
This is a measured property of these items; preserving the first parent token alone does not guarantee unchanged prefix flags.

\paragraph{Joint paths and the last flag.}
For each scored byte $x_i$ and selected prior history $h_i$, the path cost uses the joint output probability $P(x_i,b_i\mid h_i)$.
The last flag is summed out because no subsequent prediction depends on it.
The flag component at other positions is
\begin{equation}
 f_i=-\log_2 P(b_i\mid x_i,h_i)
 =\log_2\frac{P(x_i,0\mid h_i)+P(x_i,1\mid h_i)}{P(x_i,b_i\mid h_i)}.
\end{equation}
Each nonnegative $f_i$ is the surprisal of the selected flag conditional on the byte and history, rather than the entropy of the flag distribution.
Subtracting these terms leaves byte probabilities conditioned on the same selected histories.
A text likelihood would instead sum over alternative boundary histories, which can change later predictions.
This is the byte-level counterpart of summing a tokenized model over all tokenizations of a string, including noncanonical ones \citep{cao2021marginal,chirkova2023marginalize}.
The remainder therefore provides neither a corrected text likelihood nor a bound on the parent/conversion likelihood gap.

\begin{table}[htbp]
\caption{Bwen flag decomposition on the matched original-sentence subset. Parent, path, gap, flag, and remainder columns are bits per scored byte; the final column is flag share. Negative remainders do not establish a marginal-likelihood advantage.}
\label{tab:flag}
\centering\small\setlength{\tabcolsep}{4pt}
\begin{tabular}{lrrrrrr}
\toprule
 & \multicolumn{5}{c}{Bits per scored byte} &  \\
\cmidrule(lr){2-6}
Language & Parent & Bwen path & Gap & Flag term & Gap $-$ flag & Flag share \% \\
\midrule
Assamese & 0.7307 & 0.9526 & 0.2219 & 0.0855 & 0.1364 & 8.97 \\
Bengali & 0.5550 & 0.6505 & 0.0955 & 0.0488 & 0.0467 & 7.50 \\
Gujarati & 0.6361 & 0.7689 & 0.1328 & 0.0959 & 0.0369 & 12.48 \\
Hindi & 0.5150 & 0.5798 & 0.0648 & 0.0907 & \textminus{}0.0259 & 15.64 \\
Kannada & 0.5734 & 0.7487 & 0.1753 & 0.1001 & 0.0752 & 13.37 \\
Khmer & 0.5775 & 0.8574 & 0.2799 & 0.0979 & 0.1820 & 11.42 \\
Malayalam & 0.5061 & 0.6879 & 0.1818 & 0.1192 & 0.0626 & 17.33 \\
Marathi & 0.5900 & 0.6877 & 0.0977 & 0.0870 & 0.0106 & 12.66 \\
Burmese & 0.5265 & 0.7797 & 0.2532 & 0.1213 & 0.1319 & 15.56 \\
Nepali & 0.6500 & 0.7312 & 0.0812 & 0.0979 & \textminus{}0.0167 & 13.39 \\
Odia & 0.6162 & 0.7826 & 0.1664 & 0.1770 & \textminus{}0.0106 & 22.62 \\
Punjabi & 0.6237 & 0.8110 & 0.1874 & 0.0780 & 0.1094 & 9.61 \\
Shan & 0.9602 & 1.6276 & 0.6675 & 0.2507 & 0.4167 & 15.40 \\
Sinhala & 0.7747 & 0.9695 & 0.1948 & 0.0859 & 0.1088 & 8.87 \\
Tamil & 0.4823 & 0.5986 & 0.1163 & 0.0807 & 0.0356 & 13.48 \\
Telugu & 0.5722 & 0.6878 & 0.1156 & 0.0909 & 0.0248 & 13.21 \\
Thai & 0.4241 & 0.8312 & 0.4071 & 0.0796 & 0.3275 & 9.58 \\
\bottomrule
\end{tabular}

\end{table}

\paragraph{Batch-composition sensitivity.}
The measured wrapper pools up to the largest patch count in a batch, leaving surplus gathered states in rows with fewer patches.
During prefill, it shifts real patch states to the right while retaining shared position identifiers, and the scoring call supplies no attention mask for padded patches.
Batching can also change decisions near the boundary threshold.
On natural text, \KnifeEdges{} of \InteriorPositions{} Bwen interior positions lie within $\ThresholdBand$ of the threshold; this threshold-distance diagnostic does not establish a hardware-independent error bound.
Table~\ref{tab:hazard} quantifies the resulting changes under this wrapper and scoring protocol.

\begin{table}[htbp]
\caption{Absolute score changes and boundary-path changes when the same original/mark/base variants are scored together instead of alone. Sample sizes differ across checkpoints.}
\label{tab:hazard}
\centering\small
\begin{tabular}{lrrrr}
\toprule
Model & Variants & Max $|\Delta|$ (bits) & Mean $|\Delta|$ (bits) & Paths moved \\
\midrule
Bwen-8B & 306 & 12.32 & 1.54 & 10 \\
Bolmo-1B & 2,550 & 41.28 & 1.83 & 11 \\
Bolmo-7B & 2,550 & 20.88 & 1.06 & 29 \\
\bottomrule
\end{tabular}

\end{table}
\FloatBarrier

\begin{table}[htbp]
\caption{Qwen prefix sensitivity. Language rows average per-sentence inflation fractions and costs; the ratio column divides summed extra cost by summed first-token cost. The panel percentage in Section~\ref{sec:scoring} is a ratio of pooled costs, not a mean of the language percentages.}
\label{tab:bos}
\centering\small
\begin{tabular}{lrrrr}
\toprule
Language & Inflation \% & Added bits & First-token bits & Ratio \\
\midrule
Assamese & 44.7 & 146.3 & 17.3 & 8.4 \\
Bengali & 61.2 & 159.1 & 18.2 & 8.7 \\
Gujarati & 76.6 & 194.3 & 19.8 & 9.8 \\
Hindi & 83.9 & 179.0 & 19.8 & 9.0 \\
Kannada & 78.7 & 199.2 & 18.2 & 11.0 \\
Khmer & 88.4 & 325.2 & 20.1 & 16.2 \\
Malayalam & 58.9 & 134.7 & 20.8 & 6.5 \\
Marathi & 74.7 & 176.3 & 21.8 & 8.1 \\
Burmese & 83.7 & 227.3 & 16.5 & 13.7 \\
Nepali & 68.7 & 193.6 & 16.1 & 12.0 \\
Odia & 89.5 & 240.6 & 21.6 & 11.1 \\
Punjabi & 84.1 & 213.9 & 17.5 & 12.2 \\
Shan & 81.5 & 396.1 & 18.8 & 21.1 \\
Sinhala & 68.4 & 254.7 & 20.8 & 12.2 \\
Tamil & 67.7 & 175.3 & 17.6 & 9.9 \\
Telugu & 62.3 & 169.3 & 20.4 & 8.3 \\
Thai & 86.4 & 168.2 & 18.5 & 9.1 \\
\bottomrule
\end{tabular}

\end{table}

Adding the declared BOS (\texttt{<|endoftext|>}) increases the cost of the unchanged suffix targets as well as making the first token scoreable.
Table~\ref{tab:bos} quantifies this alternative conditioning convention.
\FloatBarrier

\section{Tokenizer Loading in the Second Pair}
\label{app:loading}
The dolma2 and OLMo-parent tokenizers have matching serialized vocabularies and tokenizer components after merge-format normalization, but the recorded runtime loads different pretokenizer backends.
It loads a bare byte-level pretokenizer for dolma2 while preserving the split-then-byte sequence from the OLMo model repository.
Both are fast tokenizers; the discrepancy concerns segmentation behavior, not the choice of a fast or slow implementation.

To hold bank and prefix construction fixed within the behavioral pair, both arms use the OLMo parent's loaded tokenizer, overriding the conversion's declared dolma2 tokenizer.
The instability comparison between the two conversions instead uses the default dolma2 convention.
Loading checks compare vocabularies, byte mappings, and expansion tries.
Flattening pieces into bytes removes their canonical boundaries, while suffix lookup supplies vocabulary identifiers independently of those boundaries.
Consequently, identical byte inputs and expanded identifiers can coexist with different canonical tokenizations.
The behavioral comparison is conditional on the shared OLMo-parent convention.
\FloatBarrier

\section{Comparison Pitfalls}
\label{app:pitfalls}
\paragraph{Align scored bytes and state the conditioning.}
Under the Qwen tokenizer's default settings, no BOS is added and the first token provides context.
Our protocol scores the remaining tokens and the matching byte suffix in the conversion, with no EOS target.
Prepending the declared BOS changes both the conditioning and the scored span: across \BOSItems{} sentences, the cost of the unchanged suffix targets rises by \BOSInflation\%.
This increase is \BOSMultiplier{} times the additional cost of scoring the first token.
The result demonstrates why a comparison must state its prefix convention; lower cost under one convention does not establish greater fairness.

\paragraph{Avoid unequal-patch batching in the measured wrapper.}
Strings of equal byte length can have different patch counts.
The released wrapper's pooling and padding path can then introduce surplus patch states and shifted positions without a matching padding mask, a sensitivity observed in all three wrappers under this protocol.
For Bwen-8B, batching changes costs by up to \BatchMaxBits{} bits and moves the selected boundary path in \BatchPathChanges{} of \BatchVariants{} variants.
The corresponding maxima and counts are \BatchOneBMaxBits{} bits and \BatchOneBPathChanges{}/\BatchOneBVariants{} for Bolmo-1B, and \BatchSevenBMaxBits{} bits and \BatchSevenBPathChanges{}/\BatchSevenBVariants{} for Bolmo-7B.
Standalone forwards avoid this measured dependence on batch composition.

\paragraph{Keep a path cost distinct from a text likelihood.}
The joint byte/flag score follows the conversion's selected boundary history.
On the matched subset, conditional flag surprisal accounts for \FlagLanguageMean\% of Bwen's path cost as an unweighted mean across languages (Table~\ref{tab:flag}).
This decomposition does not imply a segmentation cost unique to byte models: selecting a BPE token also selects its extent during generation.
Removing flag terms leaves byte probabilities conditioned on one history, whereas marginal text likelihood sums over alternative histories.
Appendix~\ref{app:scoring} specifies the prefix audit and final-flag marginalization.

\section{Scope of the Boundary-Target Proposal}
\label{app:proposalinterpretation}
Stage-1 distillation aligns each patch with a parent token.
Deleting boundaries merges parent units into longer patches, requiring teacher representations and decoder targets to be aligned to the merged spans.
Target construction uses full-text Unicode segmentation; P5 and the probe characterize the information available to reproduce those labels at inference.
The existing boundary loss is the natural supervision point, with the Stage-1 pipeline changes described in Sections~\ref{sec:background} and~\ref{sec:proposal}.

Boundary placement and patch count are separate experimental factors.
Isolating cluster alignment requires an unaligned control with the same per-sentence target count, together with comparisons at matched realized density.
Within an encoding, a target comparison must also hold initialization, architecture, and evaluation fixed.
Across encodings, preserving the unrestricted parent target requires an explicit remapping policy; direct transfer to UBE need not be representable.

Matching training text leaves computational cost dependent on the local encoder, global transformer, local decoder, and output head.
Target and realized $\ain$, $\bin$, patch density, edit sensitivity, and language-model quality therefore remain distinct outcomes.
Sequence lengths alone do not determine prefill or decoding latency, peak memory, or sensitivity to batch size and padding.
Language-level results are also necessary to detect regressions concealed by aggregate quality.
P1\nobreakdash--P5 characterize the target segmentation and the information available to a causal boundary predictor.
\FloatBarrier

\section{Reproducibility and Measurement Provenance}
\label{app:repro}
The public revisions, loaded-tokenizer hashes, measured parameter counts, and runtime versions below identify the artifacts used for the reported measurements.
Revision strings identify public model snapshots; digests identify the loaded artifacts and item banks.

\begin{description}
\item[\texttt{Qwen/Qwen3-8B-Base}] Public model revision \path{49e3418fbbbca6ecbdf9608b4d22e5a407081db4}; measured parameters 8,190,735,360.
\item[\texttt{allenai/Bwen-8B}] Public model revision \path{b2644f7827c21c3276d70cb82e6558aa03b9670e}; measured parameters 8,313,169,056.
\item[\texttt{allenai/OLMo-2-0425-1B}] Public model revision \path{a1847dff35000b4271fa70afc5db10fd29fedbdf}; measured parameters 1,484,916,736.
\item[\texttt{allenai/Bolmo-1B}] Public model revision \path{452690a8cfab1dd1ec27b74f2667632f17a21b46}; measured parameters 1,468,911,776.
\item[\texttt{allenai/Bolmo-7B}] Public model revision \path{11586e083d5f02977045893f3a0f38e41fba1d0a}; measured parameters 7,633,084,576.
\item[Qwen loaded tokenizer] SHA-256: \path{41e00eccf531cffc2e562d38bdd879d41e5044ea279af5b73c6a32aabcc8fe04}.
\item[OLMo loaded tokenizer] SHA-256: \path{f7499d393a5a28519bdb8de7300c517e1fb58070cfc36a25dbaecf426f824eb8}.
\item[dolma2 loaded tokenizer] SHA-256: \path{4cc03428cb83f9b7d544d07d7f984272392126854269940958df2addf38dedde}.
\item[Boundary-audit encoder] SHA-256: \path{183e49685c48270ba4b12cf84908cbae21fbdb56ca6c33b696734a93cfe9cc3f}.
\item[Qwen/Bwen probe bank] SHA-256: \path{6219dd9bc1fbd09a6a8cb07ed35ddef488163eb7bd4d93adaa4008f6bcf59cbc}.
\item[OLMo/Bolmo probe bank] SHA-256: \path{fbbd372b800a0f9cb113d068ace270114d9d10da98f114d1ba3be3528ac08a84}.
\item[dolma2 diagnostic bank] SHA-256: \path{244a573960ab3b022105ffd5a1e4deac852579d286ca81585193b02d0f1633fb}.
\item[Worked-example tokenizer] \texttt{Qwen/Qwen3-8B}, revision \path{b968826d9c46dd6066d109eabc6255188de91218}; serialized tokenizer file SHA-256: \path{aeb13307a71acd8fe81861d94ad54ab689df773318809eed3cbe794b4492dae4} (distinct from the loaded-backend serialization above).
\item[Worked-example corpus] FLORES-200 devtest (all languages), exported from the Hugging Face \texttt{facebook/flores} loader; SHA-256 of our JSON export: \path{3e2544bc200dff3d4060069be11efd9d31ddbf89101114851f7db811f553b27b}.
\item[Worked-example native tokenizer] SHA-256: \path{e01eb44f3864273391a184c2ba86879b2857e0990850a2deee98943782291d5a}.
\item[Lookahead probe] Training text \texttt{HuggingFaceFW/fineweb-2} revision \path{af9c13333eb981300149d5ca60a8e9d659b276b9}, first shard per language; evaluation on the FLORES-200 devtest cache above; runtime: torch 2.10.0+cu128; regex 2025.11.3; numpy 1.26.4; one NVIDIA GeForce RTX 3090 with deterministic algorithms.
\item[Korean/Chinese extension] The same FineWeb-2 revision and FLORES-200 cache; training sources \texttt{kor\_Hang} and \texttt{cmn\_Hani}. Runtime: torch 2.10.0+cu128; numpy 1.26.4; regex 2025.11.3; float32 on one NVIDIA GeForce RTX 3090 with deterministic algorithms. Qwen3 tokenizer revision \path{b968826d9c46dd6066d109eabc6255188de91218}; loaded-backend SHA-256: \path{41e00eccf531cffc2e562d38bdd879d41e5044ea279af5b73c6a32aabcc8fe04}.
\item[Runtime] mlstm\_kernels 2.0.6; regex 2025.11.3; tokenizers 0.22.3.dev781; torch 2.10.0+cu128; transformers 5.15.0; xlstm 2.0.4.
\item[Normalization] Unicode database 13.0.0; grapheme segmentation uses the separately versioned \texttt{regex} tables.
\end{description}

The float32 boundary audit verifies exact byte reconstruction after NFC and confirms that the unshifted offset mapping agrees with parent cuts more closely than either adjacent shift.
Its padding diagnostic verifies that changed decisions were close to the threshold before padding; a separate bfloat16 check applies autocast to the same checkpoint weights.
The cost accounting verifies exact agreement with the audit's scalar and parent-unit counts for shared languages.
Behavioral comparisons check item identities, source indices, tokenizer digests, and scored byte windows within each pair.
The flag decomposition additionally matches the subset bank and its stored path costs.

All reported corpus measurements, accounting densities, and descriptive correlations derive from fixed measurement records.
Training-mixture sizes are attributed to Bolmo's authors, and the lookahead examples are constructed encoding checks.
The tables retain the recorded confidence intervals; the leave-one-language-out analysis recomputes point estimates only.
The cost and audit records' text and parent-token counts are reproduced exactly from the pinned FLORES-200 devtest cache listed above; the normalization and grapheme tables come from distinct Unicode databases.

\end{document}